\newif\ifIsJrn
\IsJrntrue
\newif\ifIsSup
\IsSupfalse

\ifIsJrn
  \RequirePackage{fix-cm}
\documentclass[twocolumn]{svjour3}
\usepackage[english]{babel}

\usepackage[usenames,dvipsnames]{color}
\usepackage{graphicx}
\usepackage{float}
\usepackage{epsfig}
\usepackage{graphicx}
\usepackage[]{algorithm2e}

\usepackage{amsmath}
\usepackage{amssymb}

\newcommand{\ignore}[1]{}

\makeatletter

\usepackage{expl3}
\ExplSyntaxOn
\newcommand\latinabbrev[1]{
  \peek_meaning:NTF . {    #1\@}  { \peek_catcode:NTF a {      #1.\@ }    {#1.\@}}}
\ExplSyntaxOff

\def\eg{\latinabbrev{e.g}}

\def\ie{\latinabbrev{i.e}}

\newcommand{\tab}{\hspace*{2em}}

\floatstyle{ruled}
\newfloat{Equation}{tbp}{zzz}[section]

\usepackage{babel}
\journalname{ArXiv Pre-Print}

 \else
\documentclass[english,10pt,twocolumn,letterpaper]{article}
  \usepackage[T1]{fontenc}
\usepackage[latin9]{inputenc}

\usepackage[usenames,dvipsnames]{color}
\usepackage{graphicx}
\usepackage{caption}
\usepackage{float}
\usepackage{epsfig}
\usepackage{graphicx}

\usepackage{times}
\usepackage{amsmath}
\usepackage{amssymb}
\usepackage{xcite}
\usepackage{xr}
\newcommand{\ignore}[1]{}

\usepackage{babel}

 \usepackage{iccv}
 \fi

\DeclareMathOperator*{\argmin}{argmin}

\definecolor{colorXu}{RGB}{237,27,36}
\definecolor{colorKeskin}{RGB}{255,195,13}
\definecolor{colorTompson}{RGB}{1,174,240}
\definecolor{colorMelax}{RGB}{237,0,140}
\definecolor{colorNN}{RGB}{129,41,145}
\definecolor{colorPXC}{RGB}{1,168,158}
\definecolor{colorDeepSeg}{RGB}{244,111,34}
\definecolor{colorEPM}{RGB}{247,163,199}
\definecolor{colorNiTE2}{RGB}{0,166,80}
\definecolor{colorFORTH}{RGB}{254,242,0}
\definecolor{colorCascade}{RGB}{128,195,66}
\definecolor{colorLRF}{RGB}{171,163,161}
\definecolor{colorDeepPrior}{RGB}{0,0,150}
\definecolor{colorHuman}{RGB}{0,0,0}

\begin{document}

\ifIsJrn
  \newcommand{\insertFrontMatter}{
\title{Depth-based hand pose estimation: methods, data, and challenges} 
\author{James Steven Supan\v{c}i\v{c} III \and
        Gregory Rogez \and
        Yi Yang \and
        Jamie Shotton \and
        Deva Ramanan}

\institute{James Steven Supan\v{c}i\v{c} III, Gregory Rogez, Deva
  Ramanan \at University of California, Irvine \\
  \email{jsupanci@uci.edu}
  \and
  Yi Yang \at Baidu Institute of Deep Learning
  \and
  Jamie Shotton \at Microsoft Research}

\date{Received: date / Accepted: date}
}
\newcommand{\insertKeywords}{\keywords{hand pose, RGB-D sensor, datasets, benchmarking}}

\newcommand{\insertEgoPerfPlots}[0]{
  \begin{figure}
  \begin{centering}
    \textbf{UCI-EGO Test Dataset}~\cite{rogezCDC4CV}
        \par\includegraphics{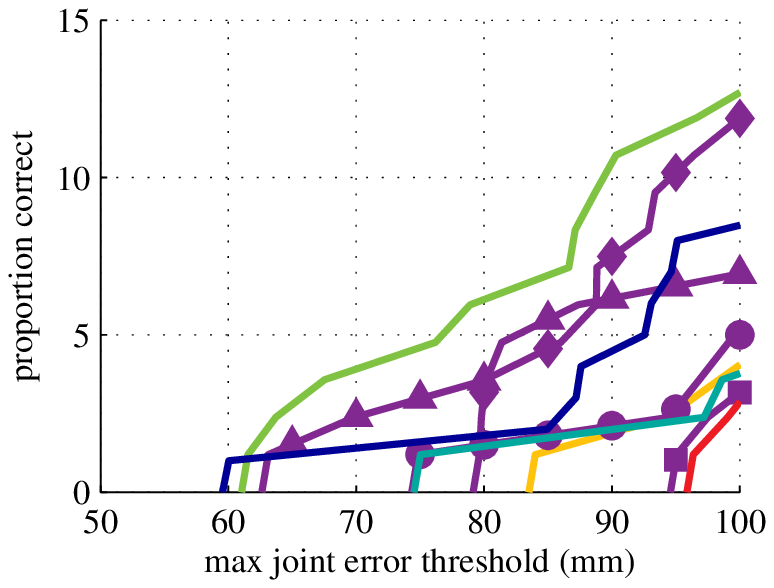}
    \par\includegraphics{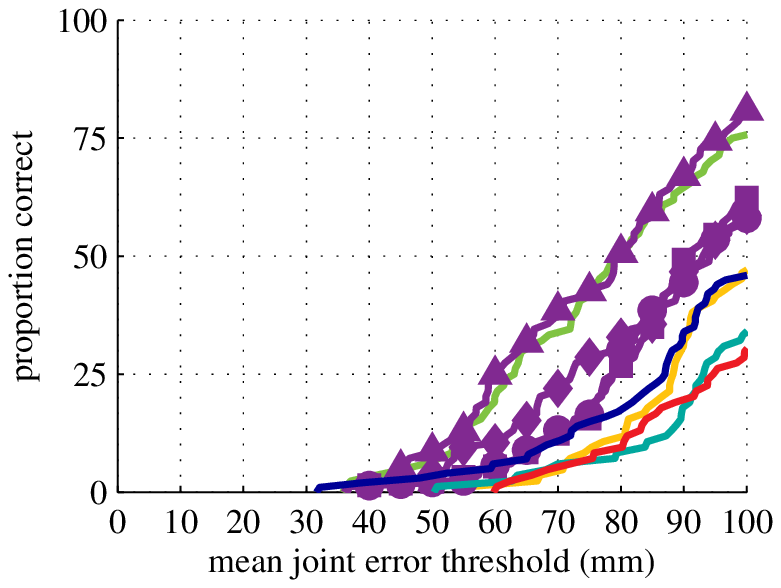}    
  \end{centering}

    \begin{minipage}{1.0\columnwidth}
    \centering
    \begin{tabular}{llll}
        \textcolor{colorNN}{NN-Ego} & \includegraphics[width=.1\columnwidth]{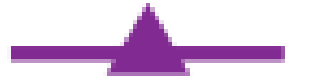} &
        \textcolor{colorNN}{NN-NYU} & \includegraphics[width=.1\columnwidth]{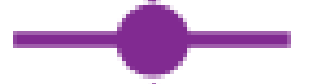} \\
        \textcolor{colorNN}{NN-ICL} & \includegraphics[width=.1\columnwidth]{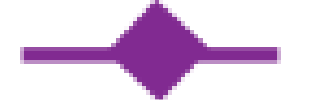} &
        \textcolor{colorNN}{NN-libhand} & \includegraphics[width=.1\columnwidth]{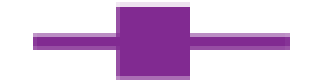} \\
        \textcolor{colorXu}{Hough~\cite{Xu2013}} &
        \textcolor{colorKeskin}{RDF~\cite{keskin2012hand}} &
        \textcolor{colorPXC}{PXC~\cite{Intel:PXC}} &        
        \textcolor{colorCascade}{R. Cascades~\cite{rogezCDC4CV}} \\
        \textcolor{colorDeepPrior}{DeepPrior~\cite{Oberweger}}
        \end{tabular}
    \end{minipage}

  \caption{\label{fig:ego-quant} For UCI-EGO, randomized cascades and our
    NN baseline do about as well, but overall, performance is
    considerably worse than other datasets. No methods are able to correctly estimate the pose (within 50mm) on {\em any} frames. Egocentric scenes contain more background clutter and object/surface interfaces, making even hand detection challenging for many methods.}
\end{figure}}

\newcommand{\inesrtNYUPerfPlots}[0]{
\begin{figure}
  \begin{centering}
    \textbf{NYU Test Dataset}~\cite{tompson14tog}
        \par\includegraphics{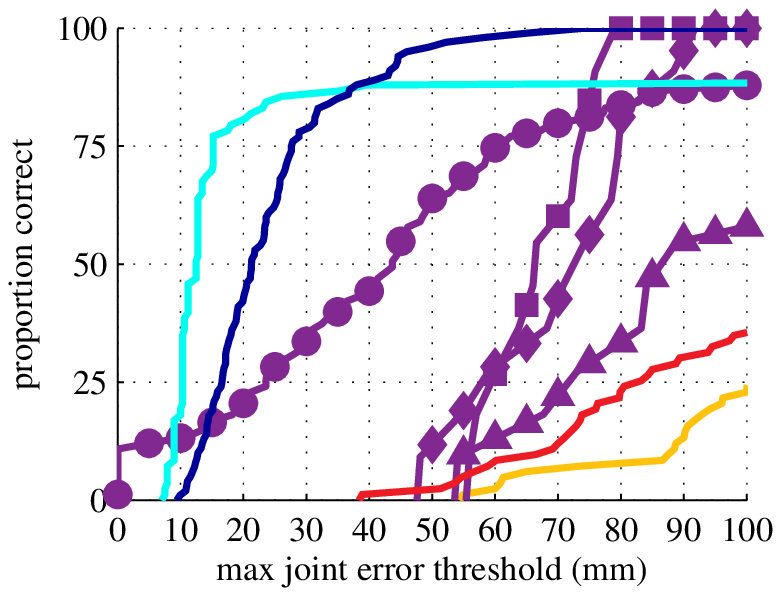}
    \par\includegraphics{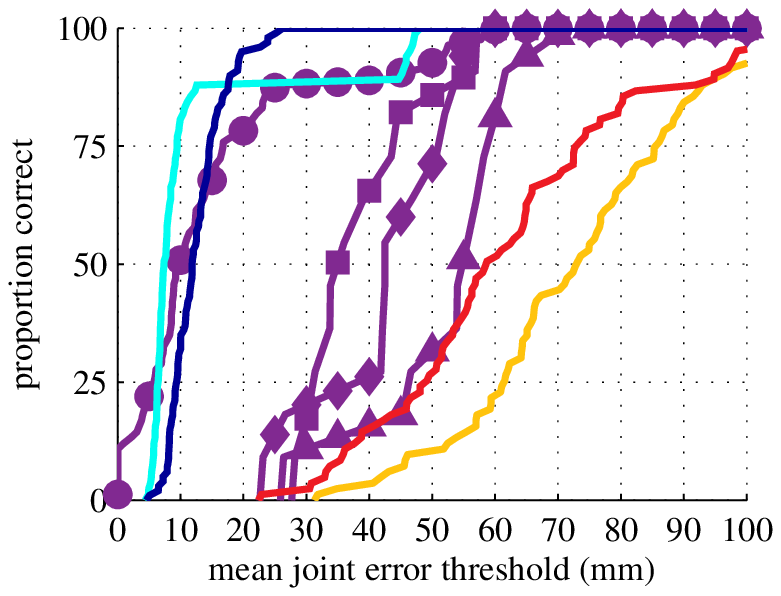}
  \end{centering}

    \begin{minipage}{1.0\columnwidth}
    \centering
    \begin{tabular}{llll}
        \textcolor{colorNN}{NN-Ego} & \includegraphics[width=.1\columnwidth]{ego-train.png} &
        \textcolor{colorNN}{NN-NYU} & \includegraphics[width=.1\columnwidth]{nyu-train.png} \\
        \textcolor{colorNN}{NN-ICL} & \includegraphics[width=.1\columnwidth]{icl-train.png} &
        \textcolor{colorNN}{NN-libhand} & \includegraphics[width=.1\columnwidth]{libhand-train.png} \\
        \textcolor{colorXu}{Hough~\cite{Xu2013}} &
        \textcolor{colorKeskin}{RDF~\cite{keskin2012hand}} &
        \textcolor{colorTompson}{DeepJoint~\cite{tompson14tog}} &
        \textcolor{colorDeepPrior}{DeepPrior~\cite{Oberweger}} 
        \end{tabular}
    \end{minipage}

    \caption{\label{fig:NYU-quant}
      Deep models~\cite{tompson14tog,Oberweger}
      perform noticeably better than other systems, and appear to
      solve both articulated pose estimation and hand detection for
      uncluttered 
      single-user scenes (common in the NYU testset). 
      However, the other systems compare more
      favorably under average error. In Fig.~\ref{fig:minvsmax}, we
      interpret this disconnect by using 1-NN to 
      show that each test hand commonly matches a training example in all but one
      finger. Please see text for further discussion.}
\end{figure}}

\newcommand{\insertICLPerfPlots}[0]{
\begin{figure}
  \begin{centering}
    \textbf{ICL Test Set}~\cite{tanglatent}
            \par\includegraphics{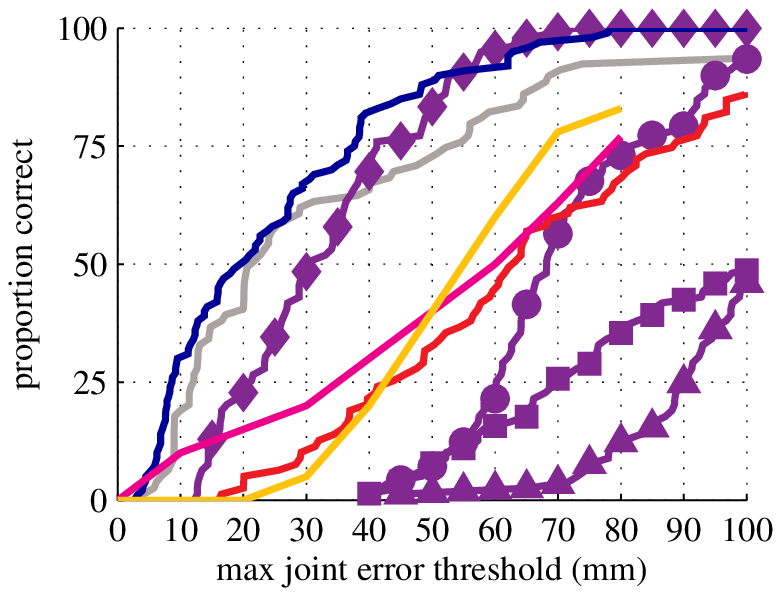}
    \par\includegraphics{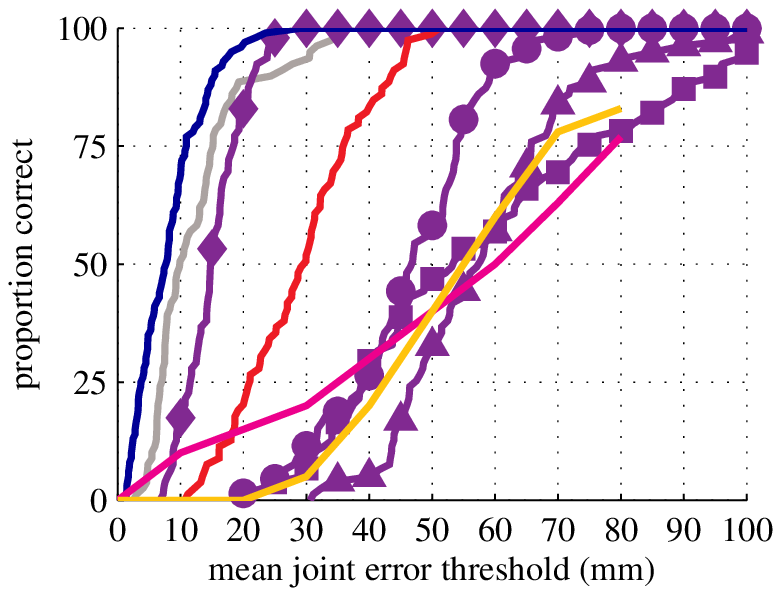}
    \end{centering}

    \begin{minipage}{1.0\columnwidth}
    \centering
    \begin{tabular}{llll}
        \textcolor{colorNN}{NN-Ego} & \includegraphics[width=.1\columnwidth]{ego-train.png} &
        \textcolor{colorNN}{NN-NYU} & \includegraphics[width=.1\columnwidth]{nyu-train.png} \\
        \textcolor{colorNN}{NN-ICL} & \includegraphics[width=.1\columnwidth]{icl-train.png} &
        \textcolor{colorNN}{NN-libhand} & \includegraphics[width=.1\columnwidth]{libhand-train.png} \\
        \textcolor{colorXu}{Hough~\cite{Xu2013}} &
        \textcolor{colorKeskin}{RDF~\cite{keskin2012hand}} &
        \textcolor{colorMelax}{Simulation~\cite{Melax2013}} &        
        \textcolor{colorDeepPrior}{DeepPrior~\cite{Oberweger}} \\
        \textcolor{colorLRF}{LRF~\cite{tanglatent}}
        \end{tabular}
    \end{minipage}
  \caption{\label{fig:ICL-quant}
  We plot results for several systems on the ICL testset using
    max-error (top) and average-error (bottom). Except for 1-NN, all systems are
    trained on the corresponding train 
    set (in this case ICL-Train). To examine cross-dataset generalization, we also plot the
    performance of our NN-baseline constructed using alternate sets
    (NYU, EGO, and libhand). When trained with ICL, NN performs as well
    or better than prior art. One can find near-perfect pose matches in
    the training set (see Fig.~\ref{fig:nn}). Please see text for
    further discussion. }

\end{figure}
}

\newcommand{\insertOurPerfPlots}[0]{
  \begin{figure*}
        \centering
    \textbf{Our Test Dataset - All Hands}
        \par
    \includegraphics{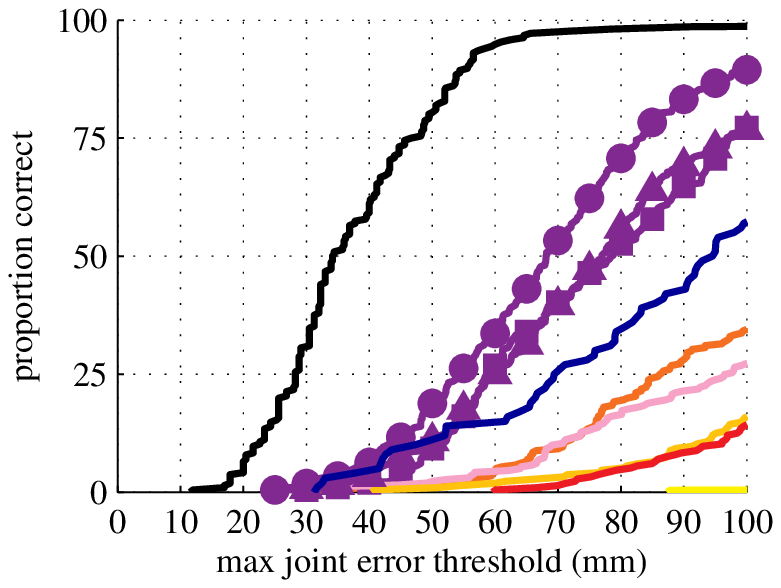}
    \includegraphics{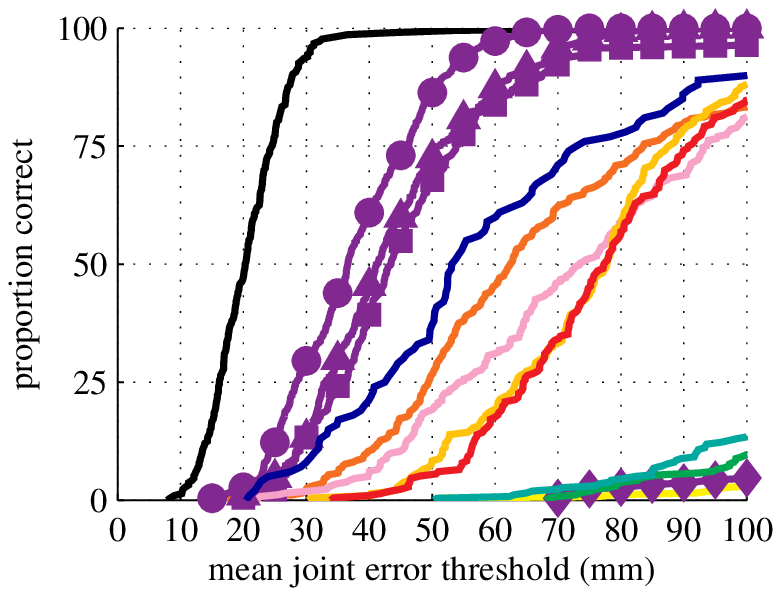}
    \par\textbf{Our Test Dataset - Near Hands ($\leq 750\text{mm}$)}
    \par
    \includegraphics{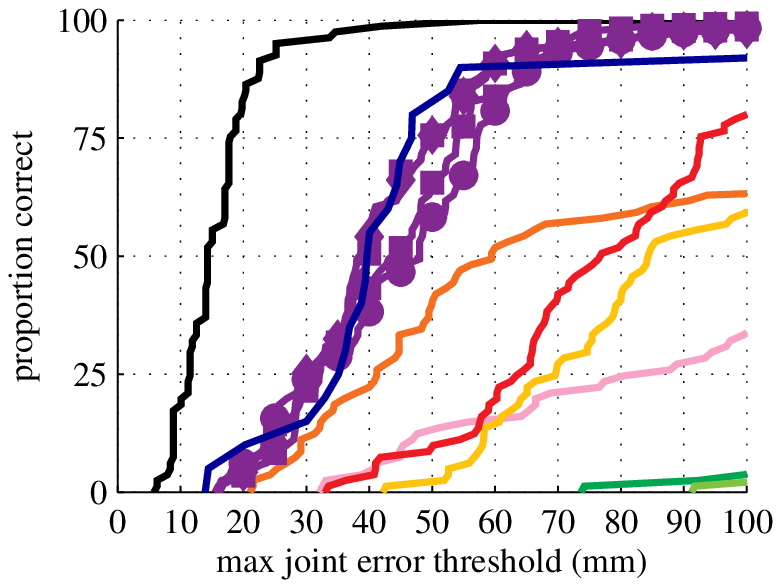}
    \includegraphics{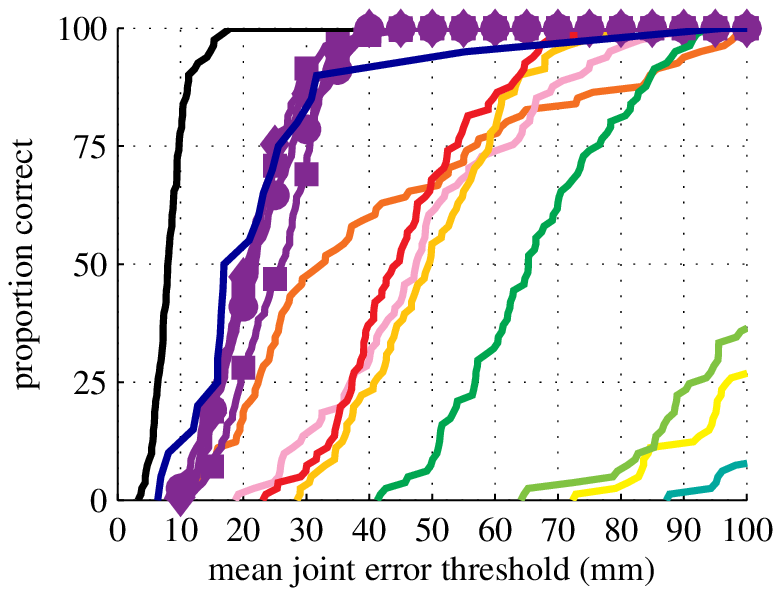}    

    \begin{minipage}{1.0\textwidth}
      \centering
    \begin{tabular}{llllllll}
        \textcolor{colorNN}{NN-Ego} \includegraphics[width=.05\columnwidth]{ego-train.png} &
        \textcolor{colorNN}{NN-NYU} \includegraphics[width=.05\columnwidth]{nyu-train.png} &
        \textcolor{colorHuman}{Human} &
        \textcolor{colorDeepPrior}{DeepPrior~\cite{Oberweger}} &
        \textcolor{colorNiTE2}{NiTE2~\cite{NiTE2}} &
        \textcolor{colorFORTH}{PSO~\cite{bmvc2011oikonom}}\\
        \textcolor{colorNN}{NN-ICL} \includegraphics[width=.05\columnwidth]{icl-train.png} &
        \textcolor{colorNN}{NN-libhand} \includegraphics[width=.05\columnwidth]{libhand-train.png}&
        \textcolor{colorEPM}{EPM}~\cite{zhu2012we}&
        \textcolor{colorDeepSeg}{DeepSeg~\cite{Couprie2013}} &
        \textcolor{colorKeskin}{RDF~\cite{keskin2012hand}} &
        \textcolor{colorPXC}{PXC~\cite{Intel:PXC}}\\
        \textcolor{colorXu}{Hough~\cite{Xu2013}} &
        \textcolor{colorCascade}{Cascades~\cite{rogezCDC4CV}} \\
        \end{tabular}      
    \end{minipage}

  \caption{\label{fig:our-quant}
    We designed our dataset to address
    the remaining challenges of in ``in-the-wild'' hand pose estimation,
    including scenes with low-res hands, clutter, object/surface interactions,
    and occlusions. We plot human-level performance (as measured through inter-annotator agreement) in black.
    On nearby hands (within 750mm, as commonly assumed in prior work)
    our annotation quality is similar to existing testsets such as
    ICL~\cite{Oberweger}. This is impressive given that our testset
    includes comparatively more ambiguous poses (see
    Sec.~\ref{subsec:disagreement}). Our dataset includes far away
    hands, for which even humans struggle to accurately
    label. Moreoever, several methods (Cascades,PXC,NiTE2,PSO) fail to
    correctly localize any hand at any distance, though the mean-error
    plots are more forgiving than the
    max-error above. In general, NN-exemplars and DeepPrior perform
    the best, correctly estimating pose on 75\% of frames with nearby
    hands. 
}
 \end{figure*}}

 \newcommand{\insertBibliography}[0]{\bibliography{hands_arxiv}}

\begin{thebibliography}{10}\itemsep=-1pt

\bibitem{Bray}
M.~Bray, E.~Koller-Meier, P.~M{\"u}ller, L.~Van~Gool, and N.~N. Schraudolph.
\newblock {3D} hand tracking by rapid stochastic gradient descent using a
  skinning model.
\newblock In {\em In 1st European Conference on Visual Media Production
  (CVMP)}, 2004.

\bibitem{Bullock2013}
I.~M. Bullock, S.~Member, J.~Z. Zheng, S.~D.~L. Rosa, C.~Guertler, and A.~M.
  Dollar.
\newblock {Grasp Frequency and Usage in Daily Household and Machine Shop
  Tasks}.
\newblock {\em Haptics, IEEE Transactions on}, 6(3):296--308, 2013.

\bibitem{Camplani2012}
M.~Camplani and L.~Salgado.
\newblock Efficient spatio-temporal hole filling strategy for kinect depth
  maps.
\newblock In {\em Proceedings of SPIE}, volume 8920, 2012.

\bibitem{castellini2011using}
C.~Castellini, T.~Tommasi, N.~Noceti, F.~Odone, and B.~Caputo.
\newblock Using object affordances to improve object recognition.
\newblock {\em Autonomous Mental Development, IEEE Transactions on},
  3(3):207--215, 2011.

\bibitem{Cooper2012}
H.~Cooper.
\newblock {Sign Language Recognition using Sub-Units}.
\newblock {\em The Journal of Machine Learning Research}, 13:2205--2231, 2012.

\bibitem{transduc:forest}
T.~Y. D.~Tang and T.-K. Kim.
\newblock Real-time articulated hand pose estimation using semi-supervised
  transductive regression forests.
\newblock In {\em ICCV}, pages 1--8, 2013.

\bibitem{Delamarre2001}
Q.~Delamarre and O.~Faugeras.
\newblock {{3D} Articulated Models and Multiview Tracking with Physical
  Forces}.
\newblock {\em Computer Vision and Image Understanding}, 81(3):328--357, Mar.
  2001.

\bibitem{deng2009imagenet}
J.~Deng, W.~Dong, R.~Socher, L.-J. Li, K.~Li, and L.~Fei-Fei.
\newblock Imagenet: A large-scale hierarchical image database.
\newblock In {\em Computer Vision and Pattern Recognition, 2009. CVPR 2009.
  IEEE Conference on}, pages 248--255. IEEE, 2009.

\bibitem{dollar2012pedestrian}
P.~Dollar, C.~Wojek, B.~Schiele, and P.~Perona.
\newblock Pedestrian detection: An evaluation of the state of the art.
\newblock {\em Pattern Analysis and Machine Intelligence, IEEE Transactions
  on}, 34(4):743--761, 2012.

\bibitem{Erol2007}
A.~Erol, G.~Bebis, M.~Nicolescu, R.~D. Boyle, and X.~Twombly.
\newblock {Vision-based hand pose estimation: A review}.
\newblock {\em Computer Vision and Image Understanding}, 108(1-2):52--73, Oct.
  2007.

\bibitem{everingham2010pascal}
M.~Everingham, L.~Van~Gool, C.~K. Williams, J.~Winn, and A.~Zisserman.
\newblock The {PASCAL} visual object classes ({VOC}) challenge.
\newblock {\em International journal of computer vision}, 88(2):303--338, 2010.

\bibitem{Couprie2013}
C.~Farabet, C.~Couprie, L.~Najman, and Y.~LeCun.
\newblock Learning hierarchical features for scene labeling.
\newblock {\em Pattern Analysis and Machine Intelligence, IEEE Transactions
  on}, 35(8):1915--1929, 2013.

\bibitem{Fei-Fei2007}
L.~Fei-Fei, R.~Fergus, and P.~Perona.
\newblock {Learning generative visual models from few training examples: An
  incremental Bayesian approach tested on 101 object categories}.
\newblock {\em Computer Vision and Image Understanding}, 106(1):59--70, Apr.
  2007.

\bibitem{feix2013}
T.~Feix, J.~Romero, C.~H. Ek, H.~Schmiedmayer, and D.~Kragic.
\newblock {A Metric for Comparing the Anthropomorphic Motion Capability of
  Artificial Hands}.
\newblock {\em Robotics, IEEE Transactions on}, 29(1):82--93, Feb. 2013.

\bibitem{felzenszwalb2010object}
P.~F. Felzenszwalb, R.~B. Girshick, D.~McAllester, and D.~Ramanan.
\newblock Object detection with discriminatively trained part-based models.
\newblock {\em Pattern Analysis and Machine Intelligence, IEEE Transactions
  on}, 32(9):1627--1645, 2010.

\bibitem{July1985}
M.~Girard and A.~A. Maciejewski.
\newblock {Computational Modeling for the Computer Animation of Legged
  Figures}.
\newblock {\em ACM SIGGRAPH Computer Graphics}, 19(3):263--270, 1985.

\bibitem{gupta2014learning}
S.~Gupta, R.~Girshick, P.~Arbel{\'a}ez, and J.~Malik.
\newblock Learning rich features from rgb-d images for object detection and
  segmentation.
\newblock In {\em Computer Vision--ECCV 2014}, pages 345--360. Springer, 2014.

\bibitem{Intel:PXC}
Intel.
\newblock Perceptual computing {SDK}, 2013.

\bibitem{janoch2013category}
A.~Janoch, S.~Karayev, Y.~Jia, J.~T. Barron, M.~Fritz, K.~Saenko, and
  T.~Darrell.
\newblock A category-level 3d object dataset: Putting the kinect to work.
\newblock In {\em Consumer Depth Cameras for Computer Vision}, pages 141--165.
  Springer London, 2013.

\bibitem{keskin2012hand}
C.~Keskin, F.~K{\i}ra{\c{c}}, Y.~E. Kara, and L.~Akarun.
\newblock Hand pose estimation and hand shape classification using
  multi-layered randomized decision forests.
\newblock In {\em ECCV 2012}, pages 852--863. 2012.

\bibitem{Li2013}
C.~Li and K.~M. Kitani.
\newblock {Pixel-Level Hand Detection in Ego-centric Videos}.
\newblock {\em 2013 IEEE Conference on Computer Vision and Pattern
  Recognition}, pages 3570--3577, June 2013.

\bibitem{martin2004learning}
D.~R. Martin, C.~C. Fowlkes, and J.~Malik.
\newblock Learning to detect natural image boundaries using local brightness,
  color, and texture cues.
\newblock {\em Pattern Analysis and Machine Intelligence, IEEE Transactions
  on}, 26(5):530--549, 2004.

\bibitem{Melax2013}
S.~Melax, L.~Keselman, and S.~Orsten.
\newblock {Dynamics based {3D} skeletal hand tracking}.
\newblock {\em Proceedings of the ACM SIGGRAPH Symposium on Interactive {3D}
  Graphics and Games - I3D '13}, page 184, 2013.

\bibitem{Moore2000}
A.~Moore, A.~Connolly, C.~Genovese, A.~Gray, L.~Grone, N.~{Kanidoris II},
  R.~Nichol, J.~Schneider, A.~Szalay, I.~Szapudi, and L.~Wasserman.
\newblock {Fast Algorithms and Efficient Statistics: N-point Correlation
  Functions}.
\newblock Dec. 2000.

\bibitem{Muja2014}
M.~Muja and D.~G. Lowe.
\newblock {Scalable Nearest Neighbor Algorithms for High Dimensional Data}.
\newblock {\em IEEE Transactions on Pattern Analysis and Machine Intelligence},
  36(11):2227--2240, Nov. 2014.

\bibitem{Oberweger}
M.~Oberweger, P.~Wohlhart, and V.~Lepetit.
\newblock {Hands Deep in Deep Learning for Hand Pose Estimation}.
\newblock {\em Computer Vision Winter Workshop (CVWW)}, 2015.

\bibitem{Ohn-Bar2014}
E.~Ohn-Bar and M.~M. Trivedi.
\newblock {Hand Gesture Recognition in Real Time for Automotive Interfaces: A
  Multimodal Vision-Based Approach and Evaluations}.
\newblock {\em IEEE Transactions on Intelligent Transportation Systems}, pages
  1--10, 2014.

\bibitem{bmvc2011oikonom}
I.~Oikonomidis, N.~Kyriazis, and A.~Argyros.
\newblock Efficient model-based {3D} tracking of hand articulations using
  kinect.
\newblock In {\em BMVC}, 2011.

\bibitem{Pang2013}
Y.~Pang and H.~Ling.
\newblock {Finding the Best from the Second Bests - Inhibiting Subjective Bias
  in Evaluation of Visual Tracking Algorithms}.
\newblock {\em 2013 IEEE International Conference on Computer Vision}, pages
  2784--2791, Dec. 2013.

\bibitem{pieropanmanip2013}
A.~Pieropan, G.~Salvi, K.~Pauwels, and H.~Kjellstrom.
\newblock Audio-visual classification and detection of human manipulation
  actions.
\newblock {\em In International Conference on Intelligent Robots and Systems},
  2014.

\bibitem{premaratne2010human}
P.~Premaratne, Q.~Nguyen, and M.~Premaratne.
\newblock {\em Human computer interaction using hand gestures}.
\newblock Springer, 2010.

\bibitem{NiTE2}
PrimeSense.
\newblock Nite2 middleware, 2013.
\newblock Version 2.2.

\bibitem{qianrealtime}
C.~Qian, X.~Sun, Y.~Wei, X.~Tang, and J.~Sun.
\newblock Realtime and robust hand tracking from depth.
\newblock {\em CVPR 2014}.

\bibitem{ren2011robust}
Z.~Ren, J.~Yuan, and Z.~Zhang.
\newblock Robust hand gesture recognition based on finger-earth mover's
  distance with a commodity depth camera.
\newblock In {\em Proceedings of the 19th ACM international conference on
  Multimedia}, pages 1093--1096. ACM, 2011.

\bibitem{rogezCDC4CV}
G.~Rogez, M.~Khademi, J.~S.~S. III, J.~M.~M. Montiel, and D.~Ramanan.
\newblock {3D} hand pose detection in egocentric rgb-d images.
\newblock {\em CDC4CV Workshop,ECCV 2014}, 2014.

\bibitem{Romero}
J.~Romero, H.~Kjellstr, and D.~Kragic.
\newblock {Monocular Real-Time {3D} Articulated Hand Pose Estimation}.
\newblock {\em Humanoid Robots, 2009. Humanoids 2009. 9th IEEE-RAS
  International Conference on}, pages 87--92, 2009.

\bibitem{russakovsky2013detecting}
O.~Russakovsky, J.~Deng, Z.~Huang, A.~C. Berg, and L.~Fei-Fei.
\newblock Detecting avocados to zucchinis: what have we done, and where are we
  going?
\newblock In {\em Computer Vision (ICCV), 2013 IEEE International Conference
  on}, pages 2064--2071. IEEE, 2013.

\bibitem{Scharstein2002}
D.~Scharstein.
\newblock {A Taxonomy and Evaluation of Dense Two-Frame Stereo}.
\newblock {\em International journal of computer vision}, 47(1):7--42, 2002.

\bibitem{shakhnarovich2003fast}
G.~Shakhnarovich, P.~Viola, and T.~Darrell.
\newblock Fast pose estimation with parameter-sensitive hashing.
\newblock In {\em Computer Vision, 2003. Proceedings. Ninth IEEE International
  Conference on}, pages 750--757. IEEE, 2003.

\bibitem{shotton2013real}
J.~Shotton, T.~Sharp, A.~Kipman, A.~Fitzgibbon, M.~Finocchio, A.~Blake,
  M.~Cook, and R.~Moore.
\newblock Real-time human pose recognition in parts from single depth images.
\newblock {\em Communications of the ACM}, 56(1):116--124, 2013.

\bibitem{Song2014}
S.~Song and J.~Xiao.
\newblock {Sliding Shapes for {3D} Object Detection in Depth Images}.
\newblock {\em ECCV 2014}, pages 634--651, 2014.

\bibitem{Sridhar}
S.~Sridhar, A.~Oulasvirta, and C.~Theobalt.
\newblock {Interactive Markerless Articulated Hand Motion Tracking Using RGB
  and Depth Data}.
\newblock {\em Computer Vision (ICCV), 2013 IEEE International Conference on}.

\bibitem{Stenger2006}
B.~Stenger, A.~Thayananthan, P.~H.~S. Torr, and R.~Cipolla.
\newblock {Model-based hand tracking using a hierarchical Bayesian filter.}
\newblock {\em IEEE transactions on pattern analysis and machine intelligence},
  28(9):1372--84, Sept. 2006.

\bibitem{stokoe2005sign}
W.~C. Stokoe.
\newblock Sign language structure: An outline of the visual communication
  systems of the american deaf.
\newblock {\em Journal of deaf studies and deaf education}, 10(1):3--37, 2005.

\bibitem{tanglatent}
D.~Tang, H.~J. Chang, A.~Tejani, and T.-K. Kim.
\newblock Latent regression forest: Structured estimation of {3D} articulated
  hand posture.
\newblock {\em Proc. of IEEE Conf. on Computer Vision and Pattern Recognition},
  2014.

\bibitem{taylor2014user}
J.~Taylor, R.~Stebbing, V.~Ramakrishna, C.~Keskin, J.~Shotton, S.~Izadi,
  A.~Hertzmann, and A.~Fitzgibbon.
\newblock User-specific hand modeling from monocular depth sequences.
\newblock In {\em Computer Vision and Pattern Recognition (CVPR), 2014 IEEE
  Conference on}, pages 644--651. IEEE, 2014.

\bibitem{tompson14tog}
J.~Tompson, M.~Stein, Y.~Lecun, and K.~Perlin.
\newblock Real-time continuous pose recovery of human hands using convolutional
  networks.
\newblock {\em ACM Transactions on Graphics}, 33, August 2014.

\bibitem{torralba2011unbiased}
A.~Torralba and A.~A. Efros.
\newblock Unbiased look at dataset bias.
\newblock In {\em Computer Vision and Pattern Recognition (CVPR), 2011 IEEE
  Conference on}, pages 1521--1528. IEEE, 2011.

\bibitem{vezhnevets2003survey}
V.~Vezhnevets, V.~Sazonov, and A.~Andreeva.
\newblock A survey on pixel-based skin color detection techniques.
\newblock In {\em Proc. Graphicon}, volume~3, pages 85--92. Moscow, Russia,
  2003.

\bibitem{libhand}
M.~\v{S}ari\'{c}.
\newblock Libhand: A library for hand articulation, 2011.
\newblock Version 0.9.

\bibitem{Xu2013}
C.~Xu and L.~Cheng.
\newblock {Efficient Hand Pose Estimation from a Single Depth Image}.
\newblock {\em 2013 IEEE International Conference on Computer Vision}, pages
  3456--3462, Dec. 2013.

\bibitem{YiYangFMP}
Y.~Yang and D.~Ramanan.
\newblock {Articulated pose estimation with flexible mixtures-of-parts}.
\newblock {\em IEEE Pattern Analysis and Machine Intelligence}, 2013.

\bibitem{zhu2012we}
X.~Zhu, C.~Vondrick, D.~Ramanan, and C.~Fowlkes.
\newblock Do we need more training data or better models for object detection?.
\newblock In {\em BMVC}, pages 1--11, 2012.

\end{thebibliography}
 \fi

\newcommand{\deva}[1]{\textcolor{blue}{[Deva: {#1}]}}
\newcommand{\james}[1]{\textcolor{Maroon}{{#1}}}

\makeatletter
\makeatother

\insertFrontMatter

\maketitle  

\begin{abstract}
Hand pose estimation has matured rapidly in recent years.
The introduction of commodity depth sensors and a multitude of practical
applications have spurred new advances. 
We provide an extensive analysis of the state-of-the-art, focusing on
hand pose estimation from a single depth frame.
 To do so, we have implemented a considerable number of systems, and will release all
software and evaluation code.
We summarize important conclusions here: (1) Pose estimation appears
roughly solved for scenes with isolated hands. However,
methods still struggle to analyze cluttered scenes where hands may be
interacting with nearby objects and surfaces. To spur further progress we introduce
a challenging new dataset with diverse, cluttered scenes. (2) Many methods evaluate themselves with disparate criteria,
making comparisons difficult. We define a consistent evaluation criteria,
rigorously motivated by human experiments. (3) We introduce
a simple nearest-neighbor baseline that
outperforms most existing systems. This implies that
 most systems do not generalize
beyond their training sets. This also reinforces the under-appreciated point that training data is
as important as the model itself. 
We conclude with directions for future progress.

\insertKeywords
\end{abstract}

\section{Introduction}
Human hand pose estimation empowers many practical applications, for example sign
language recognition~\cite{keskin2012hand}, visual 
interfaces~\cite{Melax2013}, and driver
analysis~\cite{Ohn-Bar2014}. Recently introduced consumer depth
cameras have spurred a flurry of new advances
\ifIsJrn
~\cite{qianrealtime,Melax2013,keskin2012hand,Xu2013,tanglatent,transduc:forest,Li2013,tompson14tog,Sridhar,ren2011robust}.
\fi

\paragraph{Motivation: }
Recent methods have demonstrated impressive results.
But differing (often in-house) testsets, varying performance criteria,
 and annotation errors
 impede reliable comparisons~\cite{Oberweger}.
\ifIsJrn
Indeed, a recent meta-level analysis of object tracking papers
reveals that it is difficult to trust
the ``best'' reported method in any one paper~\cite{Pang2013}.
\fi
In the field of object recognition, comprehensive benchmark evaluation has been vital for progress\ifIsJrn
~\cite{Fei-Fei2007,everingham2010pascal,deng2009imagenet}.
\fi
Our goal is to similarly diagnose the state-of-affairs, and to suggest
future strategic directions, for depth-based hand pose estimation.

\begin{figure}
   \centering
    \includegraphics[width=.95\columnwidth]{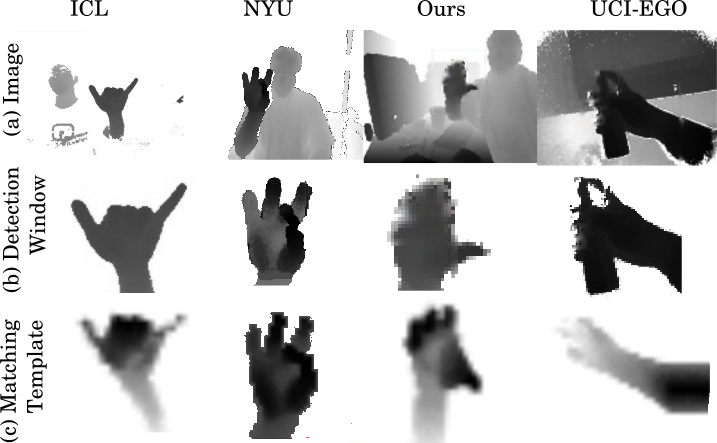}
  \caption{\label{fig:nn}
  \textbf{NN Memorization: } We evaluate a broad collection of hand pose estimation algorithms on
    different training and testsets under consistent evaluation
    criteria. Test sets which contained limited variety, in pose and
    range, or which lacked complex backgrounds were notably easier. To
    aid our analysis, we introduce a simple 3D exemplar
    (nearest-neighbor) baseline that both detects and estimates pose
    suprisingly well, outperforming most existing systems. We show the
    best-matching detection window in ({\bf b}) and the best-matching
    exemplar in ({\bf c}). We use our baseline to rank dataset
    difficulty, compare algorithms, and illustrate the importance of training set design.
      \ifIsJrn
 We provide a detailed analysis of which problem types are
    currently solved, what open research challenges remain, and
    provide suggestions for future model architectures.  
\fi
}
\end{figure}
 
\paragraph{Contributions: }
Foremost, we contribute the \emph{most extensive} 
evaluation of depth-based hand pose estimators to date. We evaluate 13
state-of-the-art hand-pose estimation systems across 4 testsets under
uniform scoring criteria. 
Additionally, we
provide a broad {\em survey} of contemporary approaches, introduce a
{\em new testset} that addresses prior limitations, and propose
a {\em new baseline} for pose estimation based on nearest-neighbor
(NN) exemplar volumes.
Surprisingly, we find that NN exceeds
the accuracy of most existing systems.
We organize our discussion 
along three axes: test data
(Sec.~\ref{sec:test}), training data (Sec.~\ref{sec:train}), and model
architectures (Sec.~\ref{sec:models}).
We survey and
taxonomize approaches for each dimension, and also contribute
novelty to each dimension (\eg \ new data and models).
After
explicitly describing our experimental protocol (Sec.~\ref{sec:eval}), we
end with an extensive empirical analysis (Sec.~\ref{sec:exp}). 

\paragraph{Preview:} We foreshadow our conclusions here. When
hands are easily segmented or detected, current systems perform quite
well. However, hand ``activities'' involving interactions with
objects/surfaces are still challenging (motivating the introduction of
our new dataset).
Moreover, in such cases even humans perform imperfectly.  
For reasonable error measures, annotators disagree 20\%
of the time (due to self and inter-object occlusions and low
resolution). This has
immediate implications for test benchmarks, but also imposes a
challenge when collecting and annotating training data. Finally, our
NN baseline illustrates some surprising points. Simple memorization of
training data performs quite well, outperforming most existing
systems. Variations in the training data often dwarf variations in the
model architectures themselves (\eg, decision forests versus deep
neural nets). Thus, our analysis offers the salient conclusion that
``it's all about the (training) data''. 

\ignore{
\begin{figure}
\begin{centering}
\includegraphics[width=.95\columnwidth]{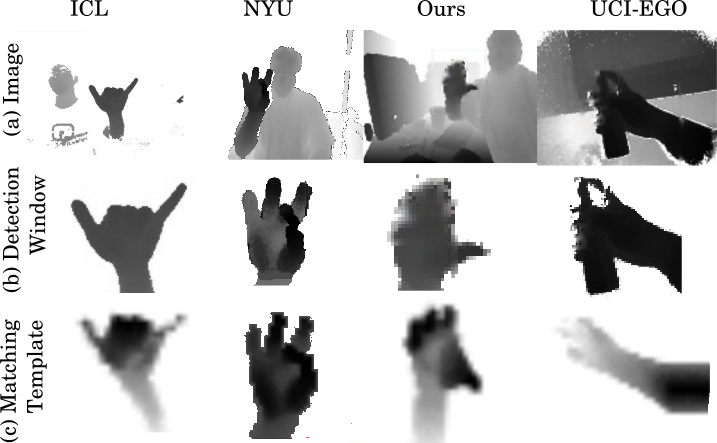} 
\end{centering}
\caption{First nearest neighbor per dataset: We examine numerous evaluation
  datasets for single frame hand pose (see \ref{table:datasets}). We
  choose several of the apparently most difficult, and commonly used
 ~\cite{tanglatent,transduc:forest,rogezCDC4CV,tompson14tog} for
  exhaustive evaluation. For each
  dataset we show a representative testing frame and the first
  nearest neighbor from the training set. In
  general, we conclude that prior datasets lack sufficient variation in
  pose: they can construct a training set such that a 1-NN classifier
  can produce comparable performance to the state-of-the-art
  classifiers, which we evaluate~\cite{Xu2013,keskin2013real,Melax2013,tanglatent,Intel:PXC,Couprie2013,NiTE2,bmvc2011oikonom,rogezCDC4CV}. This motivates our
  introduction of a new dataset, for evaluation. Still, we find that,
  when unconstrained by computation, it is possible to outperform
  state-of-the-art using a simple 1-NN classifier. With depth images,  training data is not
  only easier to synthesize, but it is also easier to augment real
  training data. } 
\label{figure:NN}
\end{figure}
}

\paragraph{Prior work:} Our work follows in the rich tradition of
\ifIsJrn
benchmarking~\cite{everingham2010pascal,russakovsky2013detecting,dollar2012pedestrian}
\fi
and taxiomatic analysis~\cite{Scharstein2002,Erol2007}. In
particular, Erol {\em et al.}~\cite{Erol2007} provided a review of
hand pose analysis in 2007. Contemporary approaches have considerably
evolved, prompted by the introduction of commodity depth cameras. We
believe the time is right for another look. We do extensive
cross-dataset analysis (by training and testing systems on different
datasets~\cite{torralba2011unbiased}).
Human-level studies in benchmark evaluation~\cite{martin2004learning}
inspired our analysis of human-performance. 
Finally, our NN-baseline is closely inspired by non-parametric
approaches to pose estimation~\cite{shakhnarovich2003fast}. In particular, we make use of volumetric depth features in a 3D scanning-window (or volume) framework, similar to~\cite{Song2014}. However, our baseline does not require SVM training or multi-cue features, making it considerably simpler to implement.
 
\section{Testing Data}
\label{sec:test}

\begin{table}[t!]
  \begin{centering}    
\resizebox{\columnwidth}{!}{  \begin{tabular}{| c | c | c | c | c | c | c | c | c | c | c | c | c | c | c | c |}
    \hline Dataset & Chal. & Scn. & Annot. & Frms. & Sub. & Cam. & Dist.~(\text{mm}) \\
    \hline 
    ASTAR~\cite{Xu2013}          & A           & 1   &  435  & 435    & 10   & ToF    & 270-580  \\
    Dexter 1~\cite{Sridhar}     & A           & 1   & 3,157 & 3,157  & 1    & Both   & 100-989   \\
    MSRA~\cite{qianrealtime}     & A           & 1   & 2,400 & 2,400  & 6   & ToF    & 339-422   \\
    ICL~\cite{tanglatent}        & A           & 1   & 1,599 & 1,599  & 1    & Struct & 200-380  \\
\hline
    FORTH~\cite{bmvc2011oikonom} & AV       & 1   &  0    & 7,148  & 5  & Struct & 200-1110   \\
    NYU~\cite{tompson14tog}      & AV       & 1   & 8,252 & 8,252  &  2  & Struct & 510-1070   \\
\hline
    \ifIsJrn KTH~\cite{pieropanmanip2013}  & AVC & 1   &  0    & 46,000 & 9  & Struct & NA       \\\fi
    UCI-EGO~\cite{rogezCDC4CV}      & AVC & 4 &  364  & 3,640  & 2   & ToF    & 200-390   \\
    Ours                        & AVC & 10+    & 23,640 & 23,640 & 10 & Both    & 200-1950 \\
    \hline
  \end{tabular}
}
        \footnotesize{Challenges (Chal.): \tab A-Articulation\tab V-Viewpoint\tab C-Clutter} 
  \end{centering}

  \caption{\textbf{Testing data sets: } We group existing benchmark
  testsets into 3 groups based on the overall {\tt challenges}
  addressed - articulation, viewpoint, and/or background clutter. We
  also tabulate the number of captured {\tt scenes}, number of {\tt
  annotated} versus {\tt total frames}, number of {\tt subjects}, {\tt
  camera} type (structured light vs time-of-flight), and {\tt
  distance} of the hand to camera. We introduce a new dataset ({\bf
  Ours}) that contains a significantly larger range of hand depths (up to 2m), more scenes (10+), more annotated frames (24K), and more subjects (10) than prior work.}
                                     \label{table:datasets}
\end{table}

\ignore{
\begin{figure}[t!]
  \begin{centering}
    Sensor types
    \par
    \begin{tabular}{c|c}
      Structured Light & Time of Flight\\
      \includegraphics[width=.45\columnwidth]{struct2.png} & \includegraphics[width=.45\columnwidth]{tof.png} 
    \end{tabular}    
  \end{centering}
  \caption{Most testsets use either structured light or time-of-flight
    (TOF) sensors (for an overview of depth sensors see \eg~\cite{Melax2013}). Systematic differences between sensors potentially complicates the use of synthetic training data (which might not exhibit the same artifacts) and the generalizability of methods (systems trained for one camera might now work for others). To obviate these complications we apply established depth image processing techniques to remove sensor specific artifacts~\cite{Camplani2012}.}
  \label{fig:sensors}
\end{figure}}

Test scenarios for depth-based hand-pose estimation have evolved
rapidly. Early work evaluated on synthetic data, while contemporary 
work almost exclusively evaluates on real data. However, because of
difficulties in manual annotation (a point that we will revisit),
evaluation was not always quantitative - instead, it has been common to
show select frames to give a qualitative sense of performance
\ifIsJrn
~\cite{Bray,Delamarre2001,bmvc2011oikonom,pieropanmanip2013}.
\fi
However, we fundamentally assume that quantitative evaluation on real
data will be vital for continued progress. 

\begin{figure}
\begin{centering}
\includegraphics[width=.95\columnwidth]{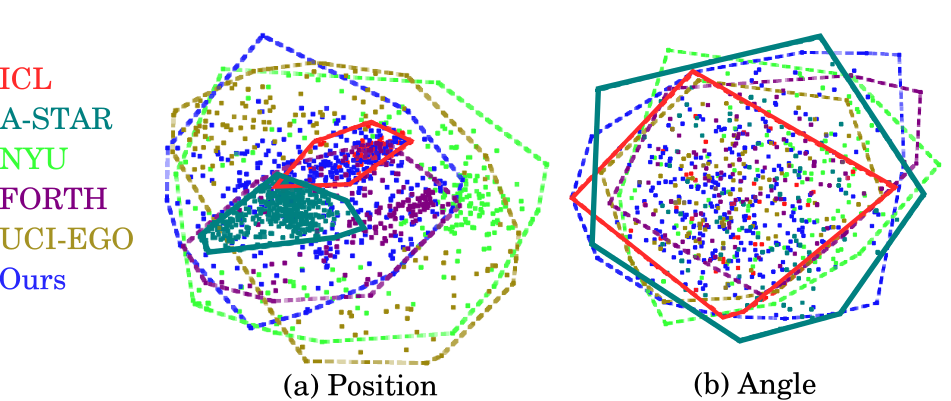} 
\par\end{centering}
\caption{\label{fig:Dataset-MDS}
\textbf{Pose variation: } We use MDS (multi-dimensional scaling) to plot the pose space covered
by various hand datasets.
For each testset, we plot the convex hull of its poses. 
We plot joint positions (left) and joint angles (right). In terms of joint angle coverage (which does not consider the ``root'' orientation of the hand itself), most datasets are similar. In terms of joint position, some datasets are limited because they consider a smaller range of viewpoints (\eg, ICL and A-STAR). We further analyze various assumptions made by datasets in the text.}
\end{figure}

\paragraph{Test set properties: } We have tabulated a list of contemporary test benchmarks in
Table~\ref{table:datasets},
\ifIsJrn
giving URLs on our website\footnote{\url{http://www.ics.uci.edu/~jsupanci/\#HandData}}.
\fi
We
refer the reader to the caption for a detailed summary of specific
dataset properties.
Per dataset, Fig.~\ref{fig:Dataset-MDS} visualizes the pose-space covered using multi-dimensional scaling (MDS).
We plot both joint 
positions (in a normalized coordinate frame that is centered and
scaled) and joint angles. Importantly, the position plot takes the
global orientation (or camera viewpoint) of the hand into account
while the angle plot does not. Most datasets are diverse in terms of
joint angles but many are limited in terms of positions (implying they
are limited in viewpoint). Indeed, we found that previous datasets
make various assumptions about articulation, viewpoint, and perhaps
most importantly, background clutter. Such assumptions are useful
because they allow researchers to focus on particular aspects of the
problem. However it is crucial to make such assumptions
explicit~\cite{torralba2011unbiased}, which much prior work does not. We do so below.

\paragraph{Articulation:} Many datasets focus on pose estimation with the assumption that detection and overall hand viewpoint is either given or limited in variation. Example datasets include MSRA~\cite{qianrealtime}, A-Star~\cite{Xu2013}, and Dexter~\cite{Sridhar}. We focus on {\bf ICL}~\cite{tanglatent} as a representative example for
 experimental evaluation because it has been used in multiple prior
 published works~\cite{tanglatent,transduc:forest}.

\paragraph{Art.~and viewpoint:} Other testsets
have focused on both viewpoint variation and articulation.
FORTH~\cite{bmvc2011oikonom} provides five test sequences with varied articulations and viewpoints, but these are unfortunately unannotated. In our experiments, we analyze the \textbf{NYU} dataset
 ~\cite{tompson14tog} because of its wider pose variation (see Fig.~\ref{fig:Dataset-MDS}) and accurate annotations (see Sec.~\ref{sec:training_data}).

\paragraph{Art.~+ View.~+ Clutter:} The most difficult datasets
contain cluttered backgrounds that are not easy to segment away. These
datasets tend to focus on ``in-the-wild'' hands undergoing activities
and interacting with nearby objects and surfaces.
\ifIsJrn
The KTH Dataset~\cite{pieropanmanip2013} provides a rich set of 3rd person videos showing humans interacting with
objects. Unfortunately, annotations are not provided for the hands
(only the objects). \fi
The \textbf{UCI-EGO}
 ~\cite{rogezCDC4CV} 
dataset provides challenging sequences from an egocentric
perspective, and so is included in our benchmark analysis.

\paragraph{Our testset:} Our empirical evaluation will show that {\em in-the-wild hand activity} is still challenging. To push research in
this direction, we have collected and annotated our own testset of real images (labeled as {\bf Ours} in  Table~\ref{table:datasets}).
As far as we are
aware, our dataset is the first to focus on hand pose estimation \emph{across multiple subjects and multiple cluttered scenes}. This is important, because any practical
application must handle diverse subjects, scenes, and clutter.

\section{Training Data}
\label{sec:training_data}

\label{sec:train}

Here we discuss various approaches for generating training data.
Real annotated training data has long been the gold standard for supervised learning. However,
 the generally accepted wisdom (for hand pose estimation) is that the
 space of poses is too large to manually annotate. This motivates
 approaches to leverage synthetically generated training data, discussed further below.

\paragraph{Real data + manual annotation:}
Arguably, the space of hand poses exceeds what can be sampled with real
data.
Our experiments identify a second
problem:  perhaps surprisingly, human annotators often disagree on
pose annotations. For example, in our testset, human annotators
visually disagreed on 20\% of pose annotations (given a
visually-acceptable threshold of 20mm) as plotted in
Fig.~\ref{fig:our-quant}. These disagreements arise from limitations in
the raw sensor data, either due to poor resolution or occlusions (as
shown in
\ifIsJrn
Sec.~\ref{subsec:disagreement}). 
\fi
These ambiguities
are often mitigated by placing the hand close to the  
camera~\cite{tanglatent,qianrealtime,Xu2013}.
As an illustrative
example, we evaluate the {\bf ICL} training set~\cite{tanglatent}.

\paragraph{Real data + automatic annotation:}
Data gloves directly obtain automatic pose annotations for real data~\cite{Xu2013}. 
However, they require painstaking per-user calibration
and distort the hand shape that is observed
in the depth map. Alternatively, one could use a ``passive'' motion
capture system. We evaluate the {\bf NYU} training
set~\cite{tompson14tog} that annotates real data by fitting (offline) a skinned 3D hand model to high-quality 3D measurements.

\paragraph{Quasi-synthetic data:}
Augmenting real data  with geometric computer graphics models 
provides an attractive solution. 
For example, one can apply geometric transformations (\eg, rotations) to both real data and its annotations~\cite{tanglatent}. If multiple depth cameras are
used to collect real data (that is then registered to a model), one
can synthesize a larger set of varied
viewpoints~\cite{tompson14tog}. Finally, mimicking the noise and
artifacts of real data is often important when using synthetic
data. Domain transfer methods~\cite{transduc:forest} learn the relationships  between a small real dataset and large synthetic one.

\paragraph{Synthetic data:} Another hope is to use data rendered by a computer graphics
system. Graphical synthesis sidesteps the annotation problem
completely: precise annotations can be rendered along with the
features. When synthesizing novel exemplars, it
is important define a good sampling distribution. The {\bf UCI-EGO} training set~\cite{rogezCDC4CV} synthesizes data with an
egocentric prior over viewpoints and grasping poses. A
common strategy for generating a sampling distribution is to collect pose samples with motion capture data~\cite{feix2013,castellini2011using}. 
\ifIsJrn
\subsection{libhand training set: }
To further examine the effect of training data, we created a massive custom training set
of 25,000,000 RGB-D training instances with the open-source {\bf libhand} model. 
We modified the code to include a forearm and output depth data,
semantic segmentations, and keypoint
annotations. We emphasize that this {\em synthetic} training
set is distinct from our new test dataset of {\em real} images.

\begin{figure}
\centering
        \includegraphics[width=.9\columnwidth]{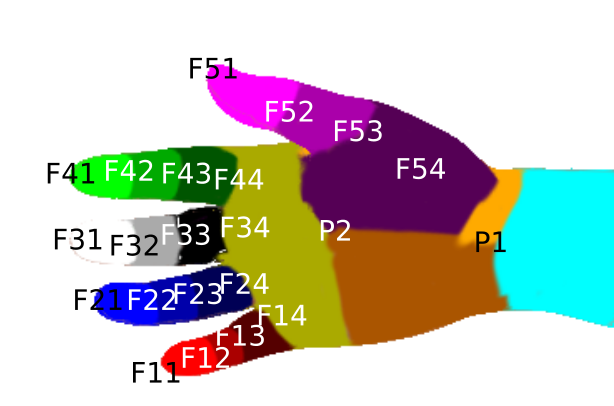} 
    \caption{\textbf{libhand joints: } We use the above joint identifiers to describe how we
    sample poses (for libhand) in table~\ref{table:synth_params}. 
      Please see \url{http://www.libhand.org/} for more details
      on the joints and their parameters.}
    \label{fig:libhand_ids}
\end{figure}

\begin{table*}
\centering
\begin{tabular}{|c|c|c|c|c|}
\hline
  Description & Identifiers & bend & side & elongation \\
\hline
  Intermediate and Distal Joints & $F_{1:4,2:3}$ & $U(\frac{-\pi}{2}^r,\frac{\pi}{7}^r)$ & 0 & 0 \\
  Proximal-Carpal Joints & $F_{1:4,4}$ & $U(\frac{-\pi}{2}^r,\frac{\pi}{7}^r)$ & $U(\frac{-\pi}{8}^r,\frac{\pi}{8}^r)$ & 0 \\
  Thumb Metacarpal & $F_{5,4}$ & $U(-1^r,.5^r)$ & $U(-.7^r,1.2^r)$ & $U(.8^r,1.2^r)$ \\
  Thumb Proximal  & $F_{5,3}$ & $U(-1^r,-.6^r)$ & $U(-.2^r,.5^r)$ & 0 \\
  Wrist Articulation & $P_1$ & $U(-1^r,1^r)$ & $U(-.5^r,.8^r)$ & 0 \\
\hline
\end{tabular}
\caption{\textbf{Synthetic hand distribution: } We render synthetic
  hands with joint angles sampled from the above
  uniform distributions. \texttt{bend} refers to the natural
  extension-retraction of the finger joints. The proximal-carpal,
  wrist and thumb joints are additionally capable
  of \texttt{side}-to-\texttt{side} articulation. We do not consider a
  third type of articulation, \texttt{twist}, because it would be
  extremely painful and result in injury. We model anatomical
  differences by \texttt{elongating} some bones fanning out from a
  joint. Additionally, we apply an isotropic global metric  scale factor sampled from the range $U(\frac{2}{3},\frac{3}{2})$. Finally, we randomize the camera viewpoint by uniformly sampling tilt, yaw and roll from $U(0,2\pi)$.  \label{table:synth_params} }
\end{table*}

\paragraph{Synthesis parameters: }
To avoid biasing our synthetic training set away from unlikely, but
possible, poses we do not use motion capture data. Instead, we take a
 brute-force approach based on rejection-sampling.
We uniformly and independently
sample joint angles (from a bounded range), and throw away invalid
samples that yield self-intersecting 3D hand poses.
Specifically, using the libhand joint identifiers shown in
figure~\ref{fig:libhand_ids}, we generate poses by uniformly sampling
from bounded ranges, as shown in Table.~\ref{table:synth_params}.

\paragraph{Quasi-Synthetic backgrounds:} An under-emphasized aspect of
synthetic training data is the choice of synthetic backgrounds. For
methods operating on pre-segmented images~\cite{keskin2012hand,qianrealtime,Sridhar}, this is likely not an issue. However, for active hands ``in-the-wild'', the choice of synthetic backgrounds, surfaces, and interacting objects is likely important. Moreover, some systems require an explicit negative set (of images not containing hands) for training. To create such a background/negative training set, we take a quasi-synthetic approach by applying random affine transformations to 5,000 images of real scenes, yielding a total of 1,000,0000 pseudo-synthetic backgrounds.  We found it useful to include human
bodies in the negative set because faces are common distractors for hand models. 

\fi

\begin{table}
  \begin{centering}    
\resizebox{\columnwidth}{!}{  \begin{tabular}{| c | c | c | c | c | c | c | c | c | c | c | c | c | c | c | c |}
    \hline Dataset & Generation & Viewpoint & Views & Size & Subj. \\
    \hline 
    ICL~\cite{tanglatent}        & Real + manual annot.             & 3rd Pers. & 1 & 331,000 & 10 \\
        \hline
    NYU~\cite{tompson14tog}      & Real + auto annot.         & 3rd Pers. & 3 & 72,757 & 1 \\
    \hline
    UCI-EGO~\cite{rogezCDC4CV}       & Synthetic                   & Egocentric& 1 & 10,000  & 1  \\
    libhand~\cite{libhand}  & Synthetic                 & Generic   & 1 & 25,000,000  & 1  \\
    \hline
  \end{tabular}
}
  \end{centering}

  \caption{\textbf{Training data sets: } We broadly categorize
  training datasets by the method used to {\tt generate} the data and annotations: real data + manual annotations, real data + automatic annotations, or synthetic data (and automatic annotations). Most existing datasets are {\tt viewpoint}-specific (tuned for 3rd-person or egocentric recognition) and limited in {\tt size} to tens of thousands of examples. 
NYU is unique in that it is a multi{\tt view} dataset collected with multiple cameras, while ICL contains shape variation due to multiple (10) {\tt subjects}. To explore the effect of training data, we use the public libhand animation package to generate a massive training set of 25 million examples.}
                               \label{table:datasets_train}
\end{table}
 
\section{Methods}

\newcommand\mcolorbox[1]{\raisebox{\height}{\colorbox{#1}{\quad}}}

\begin{table*}
  \begin{centering}  
\noindent\resizebox{\textwidth}{!}{    \begin{tabular}{|l|c|c|c|c|c|c|c|c|}
      \hline Method   & Approach & Model-drv. & Data-drv. &  Detection & Implementation & FPS \\
      \hline
      Simulate~\cite{Melax2013}                           & Tracker (simulation)          & Yes          & No     & Initialization    & Published   & $50$  \\
      NiTE2~\cite{NiTE2}                                      &Tracker (pose search)        & No           & Yes         & Initialization     & Public      & $>60$  \\
      Particle Swarm Opt. (PSO)~\cite{bmvc2011oikonom}                            & Tracker (PSO)        & Yes          & No          & Initialization      & Public&$15$   \\
      \hline 
      Hough Forest~\cite{Xu2013}   & Decision forest          & Yes          & Yes         & Decision forest & Ours       & $12$ \\
      Random Decision Forest (RDF)~\cite{keskin2012hand}          & Decision forest   & No           & Yes         & -   & Ours       & $8$  \\      
     Latent Regression Forest (LRF)~\cite{tanglatent}              & Decision forest      & No           & Yes         &-   & Published   & $62$  \\
      \hline
      DeepJoint~\cite{tompson14tog}                       & Deep network      & Yes          & Yes         & Decision forest   & Published   & $25$  \\
      DeepPrior~\cite{Oberweger} & Deep network  & No & Yes         & Scanning window       & Ours       & $5000$  \\     
      DeepSegment~\cite{Couprie2013}                         & Deep network      & No           & Yes         & Scanning window     & Ours       & $5$  \\
      \hline
      Intel PXC~\cite{Intel:PXC}                              & Morphology (convex detection)  & No           & No     &   Heuristic segment    & Public      & $>60$  \\
      Cascades~\cite{rogezCDC4CV}                  & Hierarchical cascades    & No           & Yes         & Scanning window       & Provided       & $30$  \\
      EPM~\cite{zhu2012we}    & Deformable part model       & No           & Yes         & Scanning window       & Ours       & $1/2$  \\
        Volumetric Exemplars                              & Nearest neighbor  (NN)     & No           & Yes         & Scanning volume    & Ours       & $1/15$  \\
\hline
    \end{tabular}      
}  
  \indent
  \caption{\textbf{Summary of methods: } We broadly categorize the pose estimation systems that we evaluate by their overall {\tt approach}: decision forests, deep models, trackers, or others. Though we focus on single-frame systems, we also evaluate trackers by providing them manual initialization.  {\tt Model-driven} methods make use of articulated
    geometric models at test time, while {\tt data-driven} models are
    trained beforehand on a training set. Many systems begin by {\tt
      detect}ing hands with a Hough-transform or a scanning
    window/volume search. Finally, we made use of public source code
    when available, or re-{\tt implement}ed the system ourselves,
    verifying our implementation's accuracy on published
    benchmarks. `Published' indicates that published performance
    results were used for evaluation, while `public' indicates that
    source code was available, allowing us to evaluate the method on
    additional testsets. We report the fastest speeds (in
      \texttt{FPS}), either reported or our implementation's. 
  } 
  \label{table:methods}
  \end{centering}  
\end{table*}

\label{sec:models}
Next we survey existing
approaches to hand pose estimation (summarized in Table ~\ref{table:methods}). We conclude by introducing a
simple volumetric nearest-neighbor (NN)
baseline.

\subsection{Taxonomy}

\paragraph{Trackers versus detectors:} We focus our analysis on
single-frame methods. For completeness, we also consider several tracking
baselines~\cite{bmvc2011oikonom,NiTE2,Intel:PXC} needing ground-truth
initialization. Manual initialization may provide an
unfair advantage, but we will show that single-frame methods are still
nonetheless competitive, and in most cases, outperform tracking-based
approaches. One reason is that single-frame methods essentially
``reinitialize''  themselves at each frame, while trackers cannot
recover from an error. 

\paragraph{Data-driven versus model-driven: } Historic attempts to
estimate hand pose
optimized a geometric model to fit observed data\ifIsJrn
~\cite{Delamarre2001,Bray,Stenger2006}. Recently, Oikonomidis {\em et al.}~\cite{bmvc2011oikonom} achieved
success using GPU accelerated Particle Swarm Optimization. 
\fi
However, such optimizations remain notoriously difficult due to local
minima in the objective function. As a result, model driven
systems have found their successes mostly to the tracking domain, where
initialization constrains the search space  
\cite{Sridhar,Melax2013,qianrealtime}.
For single image detection,
various fast classifiers 
\cite{keskin2012hand,Intel:PXC} have obtained real-time speeds. Most of the
systems we evaluate fall into this category. When these classifiers
are trained with data synthesized from a geometric model, they can be
seen as efficiently approximating model fitting.

\paragraph{Multi-stage pipelines:} It is common to treat the initial
detection (candidate generation) stage as separate from hand-pose estimation. Some systems
 use  special purpose detectors as a ``pre-processing''
stage
\ifIsJrn
~\cite{Xu2013,keskin2012hand,tompson14tog,Intel:PXC,bmvc2011oikonom,July1985,Cooper2012,Romero}.
\fi
Others use a geometric model for
inverse-kinematic (IK) refinement/validation during a ``post-processing'' stage~\cite{Xu2013,tompson14tog,Melax2013,Sridhar}.
A segmentation pre-processing stage has been
historically popular. Typically, the depth image is segmented with simple morphological operations~\cite{premaratne2010human} or the RGB image is segmented with skin classifiers~\cite{vezhnevets2003survey}.
\ifIsJrn
allowing features such as Zernike moments~\cite{Cooper2012} or skeletonizations~\cite{premaratne2010human} to be computed.
\fi 
The latter appears difficult to generalize across subjects and scenes with varying lighting ~\cite{qianrealtime}. We evaluate a depth-based segmentation system~\cite{Intel:PXC} for completeness. 

\subsection{Architectures}
In this section, we describe popular architectures for hand-pose estimation, placing in bold those systems that we empirically evaluate.

\paragraph{Decision forests:} Decision forests constitute a
dominant paradigm for estimating hand pose from depth. {\bf  Hough
  Forests}~\cite{Xu2013} take a two-stage approach of hand detection
followed by pose estimation. \textbf{Random Decision Forests
  (RDFs)}~\cite{keskin2012hand} and \textbf{Latent Regression Forests
  (LRFs)}~\cite{tanglatent} leave the initial detection stage
unspecified, but both make use of coarse-to-fine decision trees that
perform rough viewpoint classification followed by detailed pose
estimation. We experimented with several detection front-ends for RDFs
and LRFs, finally selecting the first-stage detector from Hough Forests for its strong performance.

\paragraph{Part model:} Pictorial structure models have
been popular in human body pose
\ifIsJrn
estimation~\cite{YiYangFMP},
\fi but they appear rare in hand pose estimation. For completeness, we evaluate a deformable part model defined on depth image
\ifIsJrn
patches ~\cite{felzenszwalb2010object}.
\fi
We specifically train an \textbf{exemplar part model (EPM)} constrained to
model deformations consistent with 3D exemplars~\cite{zhu2012we}, which will be described further in a tech report. 
\paragraph{Deep models:}
Recent systems have explored the use of deep neural nets for hand pose estimation.
We consider three variants in our experiments. \textbf{DeepJoint}~\cite{tompson14tog} uses a three stage pipeline that initially detects hands with a decision forest, regresses joint locations with a deep network, and finally refines joint predictions with inverse kinematics (IK). {\bf DeepPrior}~\cite{Oberweger} is based on a similar deep network, but does not require an IK stage and instead relies on the network itself to learn a spatial prior. \textbf{DeepSeg}~\cite{Couprie2013} takes a pixel-labeling approach, predicting joint labels for each pixel, followed by a clustering stage to produce
joint locations. This procedure is reminiscent of pixel-level part
classification of Kinect~\cite{shotton2013real}, but substitutes a deep network for
a decision forest. 

\subsection{Volumetric exemplars}
\label{subsec:NNBaseline}
We propose a nearest-neighbor (NN) baseline for additional diagnostic analysis. Specifically, we convert depth map measurements into a 3D voxel grid, and simultaneously detect and estimate pose by scanning over this grid with volumetric exemplar templates.

\paragraph{Voxel grid: } Depth cameras report depth as a
function of pixel $(u,v)$ coordinates: $D(u,v)$. To construct a voxel grid, we first re-project these image measurements into 3D using known
camera intrinsics $f_u,f_v$.  
\begin{align}
                        \left(x,y,z\right) = \left(
        \frac{u}{f_u} D(u,v),
        \frac{v}{f_v} D(u,v),
        D(u,v)\right)
                \label{eq:persp_2_ortho}
\end{align}
Given a test depth image, we construct a binary voxel grid $V[x,y,z]$ that is `1' if a depth value is observed at a quantized $(x,y,z)$ location. To cover the rough viewable region of a camera, we define a coordinate frame of $M^3$ voxels, where $M=200$ and  each voxel spans $10 \text{mm}^3$. We similarly convert training examples into volumetric exemplars $E[x,y,z]$, but instead use a smaller $N^3$ grid of voxels (where $N=30$), consistent with the size of a hand. 

\noindent{\bf Occlusions: } When a depth measurement is observed at a position $(x',y',z')=1$, all voxels behind it are occluded $z > z'$. We define occluded voxels to be `1' for both the test-time volume $V$ and training exemplar $E$. 
\paragraph{Distance measure:} Let $V_j$ be the $j^{\text{th}}$ subvolume (of size $N^3$) extracted from $V$, and let $E_i$ be the $i^{\text{th}}$ exemplar. We simultaneously detect and estimate pose by computing the best match in terms of Hamming distance:
\begin{align}
  (i^*,j^*) &=\argmin_{i,j} \text{Dist}(E_i,V_j) \qquad \text{where} \\
   \text{Dist}(E_i,V_j) &=  \sum_{x,y,z} \mathcal{I} (E_i[x,y,z] \neq V_j[x,y,z]), \label{eq:NNeq}
\end{align}
such that $i^*$ is the best-matching training exemplar and $j^*$ is its detected position.

\begin{figure}
\centering
\begin{minipage}[c]{0.4\columnwidth}
  \includegraphics[width=\columnwidth]{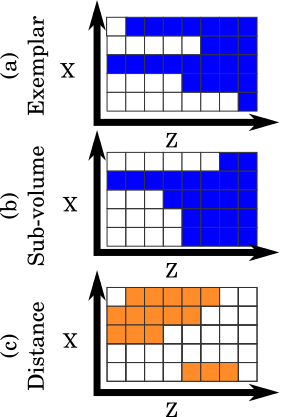}
\end{minipage}\hfill
\begin{minipage}[c]{0.55\columnwidth}
  \caption{\textbf{Volumetric Hamming distance: }We visualize 3D voxels corresponding to an exemplar (a) and
  subvolume (b). For simplicity, we visualize a 2D slice along a fixed y-value. Because occluded voxels are defined to be `1'
  (indicating they are occupied, shown in blue) the
  total Hamming distance is readily computed by the L1 distance
  between projections along the z-axis (c), mathematically shown
  in Eq.\eqref{eq:l1}. 
  } \label{figure:l1eqhamm}
\end{minipage}

\end{figure}

\paragraph{Efficient search:} A naive search over exemplars and
subvolumes is prohibitively slow. But because the underlying features
are binary and sparse, there exist considerable opportunities for
speedup. We outline two simple strategies. First, one can eliminate
subvolumes that are empty, fully occluded, or out of the camera's
field-of-view. Song {\em et al.}~\cite{Song2014} refer to such 
pruning strategies as ``jumping window'' searches. Second, one can compute volumetric Hamming distances with 2D computations:
\begin{align}
&\text{Dist}(E_i,V_j) =\sum_{x,y}\left| e_i[x,y] - v_j[x,y] \right|\ \qquad \text{where} \label{eq:l1}\\
&e_i[x,y] = \sum_z E_i[x,y,z], \qquad v_j[x,y] = \sum_z V_j[x,y,z]. \nonumber
\end{align}
Intuitively, because our 3D volumes are projections of 2.5D
measurements, they can be sparsely encoded with a 2D array (see
Fig.~\ref{figure:l1eqhamm}).
Taken together, our two simple strategies
imply that a 3D volumetric search can be made as practically efficient
as a 2D scanning-window search. For a modest number of exemplars, our
implementation still took tens of seconds per frame, which sufficed
for our offline analysis. We posit
faster NN algorithms could produce real-time performance
\ifIsJrn
~\cite{Moore2000,Muja2014}.
\fi

\begin{figure}
\centering
\begin{minipage}[c]{0.7\columnwidth}
  \caption{\textbf{Windows v. volumes: } 2D scanning windows {\bf (a)} versus 3D scanning volumes {\bf (b)}. Volumes can ignore
background clutter that lie outside the 3D scanning volume but still fall
inside its 2D projection. For example, when scoring the above hand, a
3D scanning volume will ignore depth measurements from the shoulder and head,
unlike a 2D scanning window.}
\label{fig:volume}
\end{minipage}
\begin{minipage}[c]{0.20\columnwidth}
\centering
\begin{tabular}{cc}
  (a)
  & \raisebox{-.5\height}{\includegraphics
    [width=\columnwidth,trim=50 0 0 0,clip]
        {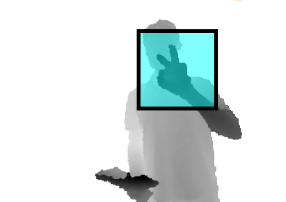}}\\
  (b) & \raisebox{-.5\height}{\includegraphics[width=\columnwidth]{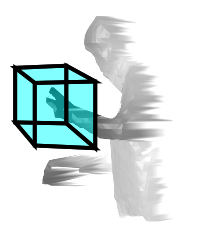}}
\end{tabular}
\end{minipage}
\vspace{-5pt}
\end{figure}

\paragraph{Comparison: } Our volumetric exemplar baseline uses a scanning volume search and 2D depth encodings. It is useful to contrast this with a ``standard'' 2D scanning-window template on depth features~\cite{janoch2013category}. First, our exemplars are defined in metric coordinates (Eq.~\ref{eq:persp_2_ortho}). This means that they will {\em not} fire on the small hands of a toy figurine, unlike a scanning window search over scales. Second, our volumetric search ensures that the depth encoding from a local window contain features only within a fixed $N^3$ volume. This gives it the ability to segment out background clutter, unlike a 2D window (Fig.~\ref{fig:volume}). 

\ifIsJrn
\section{Protocols}
\fi
\label{sec:eval}

\ifIsJrn
\subsection{Evaluation}
\fi
\begin{figure}
\centering  \includegraphics[width=.8\columnwidth]{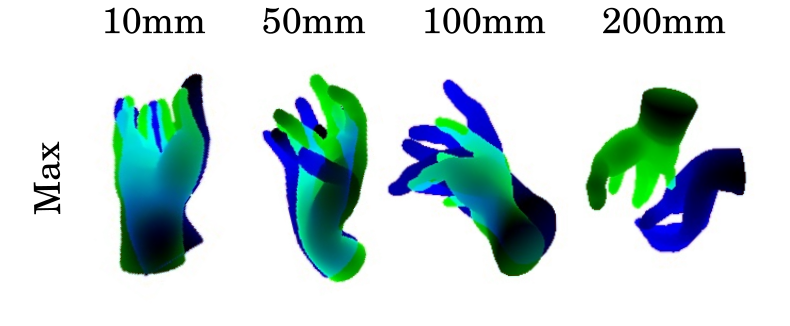}
  \caption{\label{fig:thresholds} \textbf{Our error criteria:} For each predicted hand, we
    calculate the average and maximum distance (in mm) between its
    skeletal joints and a ground-truth. In our experimental results, we plot the fraction of
    predictions that lie within a distance threshold, for various
    thresholds. This figure visually illustrates the
    misalignment associated with various thresholds for max error. A
    50mm max-error seems visually consistent with a ``roughly correct pose
    estimation'', and a 100mm max-error is consistent with a ``correct hand detection''.}
\vspace{-5pt}
\end{figure}

\paragraph{Reprojection error:} Following past work, we evaluate
pose estimation as a regression task that predicts a set of 3D joint
locations~\cite{tanglatent,bmvc2011oikonom,qianrealtime,taylor2014user,keskin2012hand}. Given
a predicted and ground-truth pose, we compute both the average and max
3D reprojection error (in mm) across all joints. We use the skeletal
joints defined by libhand~\cite{libhand}. We then summarize
performance by plotting the proportion of test frames whose average
(or max) error falls below a threshold. 

\paragraph{Error thresholds:} 
Much past work considers performance at fairly low error thresholds,
approaching 10mm~\cite{tanglatent,tompson14tog,Xu2013}.
Interestingly, \cite{Oberweger} show that established benchmarks such
as the {\bf ICL} testset include annotation errors of above 10mm in
over a third of their frames.
Ambiguities arise from manual labeling of joints versus bones and centroids versus surface points. We rigorously evaluate human-level performance through inter-annotator agreement on our new testset (Fig.~\ref{fig:our-quant}). Overall, we find that max-errors of \textbf{20mm} approach the limit of human accuracy for closeby hands. We present a qualitative visualization of max error at different thresholds in Fig.~\ref{fig:thresholds}.
{\bf 50mm} appears consistent with a roughly correct pose, while an
error within {\bf 100mm} appears consistent with a correct detection.
Our qualitative analysis is consistent with empirical studies of
human grasp~\cite{Bullock2013} and gesture~\cite{stokoe2005sign}
which also suggest that 50mm is sufficient
to capture difference in gesture or grasp. For completeness, we plot results across a
large range of thresholds, but highlight 50 and 100mm thresholds for additional analysis.

\ifIsJrn
\begin{algorithm}
  \SetKwInOut{Input}{input} \SetKwInOut{Output}{output}
  \Input{predictions and ground truths for each image}
  \Output{a set of errors, one per frame}
  \ForAll{test\_images}{
    $P \gets \text{method's most confident prediction}$\;
    $G \gets \text{ground truths for the current test\_image}$\;
    \eIf{$G = \emptyset$}{
      \tcc{Test Image contains zero hands}
      \eIf{$P = \emptyset$}{
        $\text{errors} \gets \text{errors} \cup \{0\}$\;
      }{
        $\text{errors} \gets \text{errors} \cup \{\infty\}$\;
      }
    }{
      \tcc{Test Image contains hand(s)}
      \eIf{$P = \emptyset$}{
        $\text{errors} \gets \text{errors} \cup \{\infty\}$\;
      }{
        $\text{best\_error} \gets \infty$\;
        \tcc{Find the ground truth best matching the method's prediction}
        \ForAll{$H \in G$}{
          \tcc{For mean error plots, replace $\text{max}_i$ with $\text{mean}_i$}
          \tcc{$V$ denotes the set of visible joints}
          $\text{current\_error} \gets \max_{i \in V}||H_i - P_i||_2$\;
          \If{$\text{current\_error} < \text{best\_error}$}{
            $\text{best\_error} \gets \text{current\_error}$\;
          }
        }
        $\text{errors} \gets \text{errors} \cup \{\text{best\_error}\}$\;
      }
    }
  }
\label{alg:scoring}
\caption{\textbf{Scoring Procedure: } For each frame we compute a
  max or mean re-projection error for the ground truth(s) $G$ and
  prediction(s) $P$. We later plot the proportion of frames with an
  error below a threshold, for various thresholds. }
\end{algorithm}
\fi

\paragraph{Detection issues: } Reprojection error is hard to
define during detection failures: that is, false positive hand detections
or missed hand detections. Such failures are likely in cluttered
scenes or when considering scenes containing zero or two hands. 
If a method produced zero detections when a hand was present, or
produced one if no hand was present, this was
treated as a ``maxed-out'' reprojection error (of $\infty\ \text{mm}$). If two hands
were present, we scored each method against both and took the minimum error. 
\ifIsJrn
Though we plan to release our evaluation software, we give pseudocode
in Alg.~\ref{alg:scoring}.
\fi

\paragraph{Missing data:} Another challenge with reprojection
error is missing data. First, some methods predict 2D rather than 3D joints~\cite{Intel:PXC,premaratne2010human,tompson14tog,Couprie2013}. 
Inferring depth should in theory be straightforward with Eq.~\ref{eq:persp_2_ortho}, but small 2D errors in the estimated joint can cause significant errors in the estimated depth. We report back the centroid depth of a segmented/detected hand if the measured depth lies outside the segmented volume. Past comparisons appear not to do this~\cite{Oberweger}, somewhat unfairly penalizing 2D approaches~\cite{tompson14tog}.
Second, some
methods may predict a subset of
joints~\cite{Intel:PXC,premaratne2010human}. To ensure a consistent
comparison, we force such methods to predict the locations of visible
joints with a post-processing inverse-kinematics (IK) stage~\cite{tompson14tog}. We fit
the libhand kinematic model to the predicted joints, and
infer the location of missing ones. 
Third, ground-truth joints may be occluded. By convention, we
only evaluate visible joints in our benchmark analysis.

\paragraph{Implementations:} We use 
public code when available
\cite{bmvc2011oikonom,NiTE2,Intel:PXC}. Some authors responded to our
request for their code~\cite{rogezCDC4CV}. When software was not
 available, we attempted to re-implement methods
ourselves. We were able to successfully
reimplement~\cite{Oberweger,Xu2013,keskin2012hand}, matching the accuracy on
published results~\cite{tanglatent,Oberweger}. In other cases, our in-house
implementations did not suffice~\cite{tompson14tog,tanglatent}. For these latter cases, we include published performance reports, but
unfortunately, they are limited to their own datasets. This partly
motivated us to perform a multi-dataset analysis. In particular, previous benchmarks
have shown that one can still compare algorithms across datasets using
head-to-head matchups (similar to approaches used to rank sports teams that
do not directly compete~\cite{Pang2013}). We use our NN baseline to do
precisely this. Finally, to spur further progress, {\em we will make all
implementations publicly available, together with our evaluation code.} 

\ifIsJrn
\subsection{Annotation}
\label{subsec:disagreement}
We now describe how we collect ground truth annotations. We present the annotator
 with cropped RGB and Depth images. They then
 click semantic key-points, corresponding to specific
joints, on either the RGB or Depth images. To ease the annotator's task
and to get 3D keypoints from 2D clicks we invert the forward rendering (graphics)
hand model provided by libhand which projects model parameters
$\theta$ to 2D keypoints $P(\theta)$. 
While they label
joints, an inverse kinematic solver minimizes the distance between the
currently annotated 2D joint labels, $\forall_{j \in J}L_j$, and those projected from the
libhand model parameters, $\forall_{j \in J}P_j(\theta)$.
\begin{equation}
  \label{eq:IK}  
  \min_\theta \sum_{j \in J} \lVert L_j - P_j(\theta) \rVert_2
\end{equation}
The currently fitted
libhand model, shown to the annotator,
updates online as more joints are labeled. When the
annotator indicates satisfaction with the fitted model, we proceed 
to the next frame. We give an example of the annotation process in
figure~\ref{fig:annotationinaction}.

\textbf{Strengths: } Our annotation process has several
strengths. First, kinematic constraints prevent some possible
combination of keypoints: so it is often possible to fit the model
by labeling only a subset of keypoints.
Second, the fitted
model provides annotations for occluded keypoints. Third and most
importantly, the fitted model provides 3D (x,y,z) keypoint
locations given only 2D (u,v) annotations. 

\textbf{Disagreements: } As shown in in Fig.~\ref{fig:our-quant}, annotators disagree substantially on
the hand pose, in a surprising number of cases.
In applications, such as sign
language~\cite{stokoe2005sign} ambiguous poses are typically
avoided. We believe it is important to acknowledge that, in
general, it may not be possible to achieve full precision.
Figure~\ref{fig:annotator_errors}
illustrates two examples of these annotator disagreements.

\begin{figure}
  \centering
  \ifIsSup
  \includegraphics[width=.4\columnwidth]{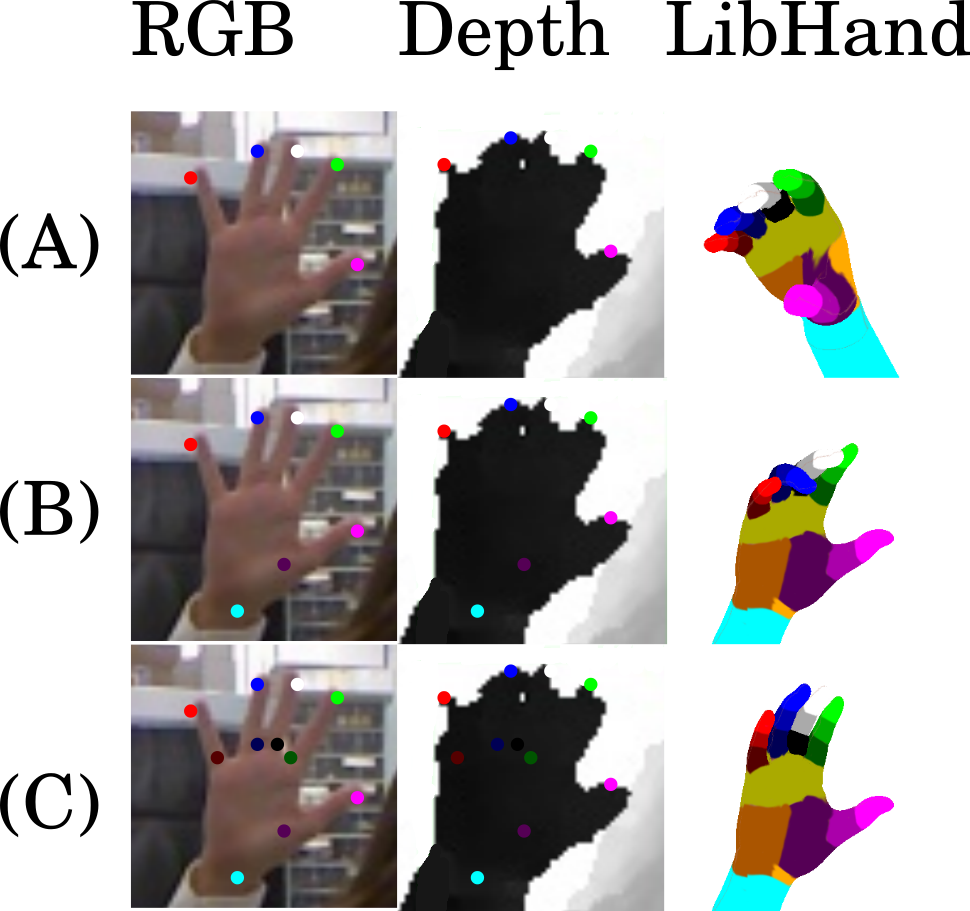}
  \else
  \ifIsJrn
  \includegraphics[width=.7\columnwidth]{annotation_process.png}
  \else
  \includegraphics[width=.4\columnwidth]{../annotation_process.png}
  \fi
  \fi
  \caption
      {\textbf{Annotation procedure: }
        We annotate until we are satisfied that
         the fitted hand pose matches the RGB and Depth data. The
         first two columns show the image evidence presented and
         keypoints received. The right most column shows the fitted
         libhand model. (A) the IK solver is able to easily
         fit a model to the five given keypoints, but it doesn't match
         the image well. (B) The annotator attempts to correct the
         model, to better match the image, by labeling the wrist. (C)
         Labeling additional finger joints finally yields and acceptable
         solution.
         \label{fig:annotationinaction}
       }
\end{figure}

\ifIsJrn
\begin{figure*}
  \else
  \begin{figure}[H]
  \fi
  \centering
  \ifIsJrn
  \includegraphics[width=.8\linewidth,clip=true]{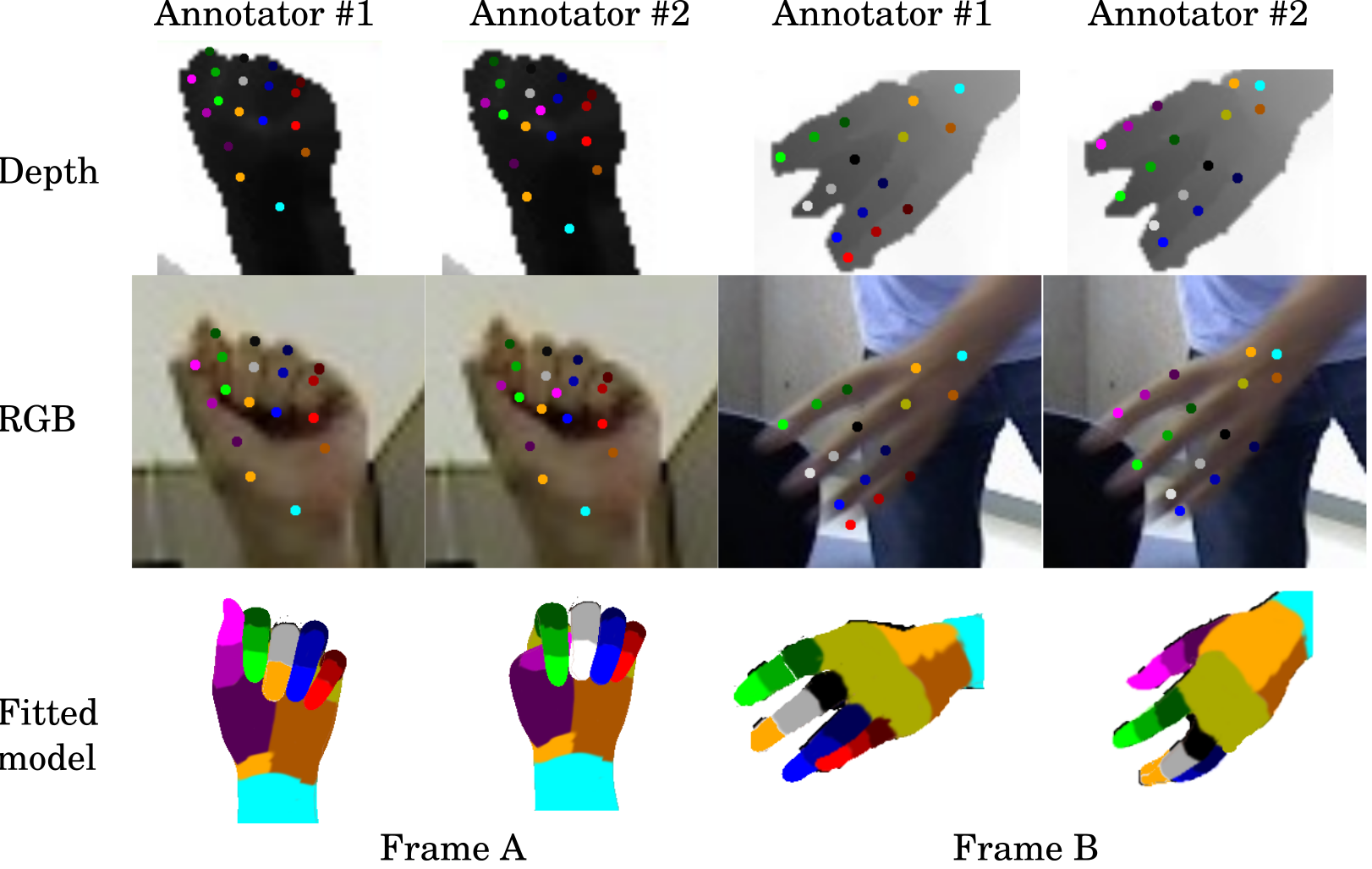}
  \else
  \includegraphics[width=.8\columnwidth,clip=true]{../disagreements.png}
  \fi
\caption[LoF entry]
    { \textbf{Annotator disagreements:} With whom do you agree? We show two frames where annotators
      disagreed. The top
      two rows show the RGB and depth images presented and the keypoint
      annotations received from the annotator. The bottom row shows the
      \texttt{libhand} model fitted to those keypoint annotations.

      \quad In \emph{Frame A}, the confusion revolves about the thumb
      position. Is the thumb occluded, folded down behind the other
      digits, or does it stand upright? The resolution, in both color
      and depth makes this hard to decide. Long range (low resolution)
      scenarios are important; But in these scenarios we cannot
      expect performance comparable to that found in near range. 
      
      \quad Similarly, in \emph{Frame B} one finger is occluded, but which one? Annotator
      1 believes the thumb is occluded. Annotator 2 believes the
      pinky is occluded. The fitted \texttt{libhand} models show that
      either interpretation is plausible. In this author's opinion,
      annotator 1 is more consistent with the RGB evidence while
      annotator 2 is more consistent with the Depth evidence. 
    }
    
    \label{fig:annotator_errors}
    \ifIsJrn
  \end{figure*}
  \else
\end{figure}
\fi
 \else
\fi
 
\section{Results}
\label{sec:results}
\label{sec:exp}

\begin{table}[]
\begin{centering}
\includegraphics[width=.95\columnwidth]{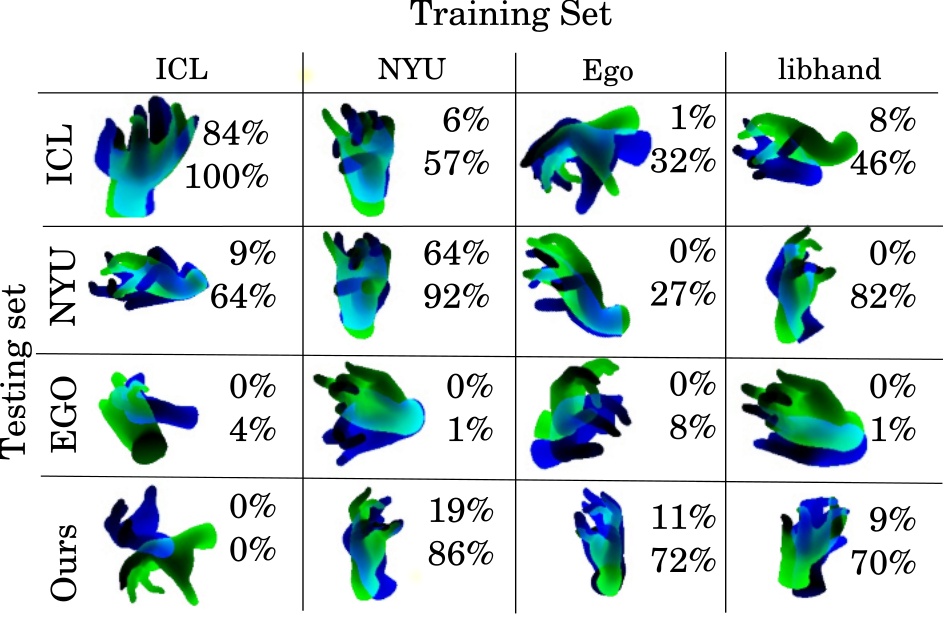} 
\end{centering}
\caption
    { \textbf{Cross-dataset generalization: }
                  We compare training and test sets using a 1-NN classifier.
    Diagonal entries represent the
    performance using corresponding train and test sets.       
                  In each grid entry, we
      denote the percentage of test frames that are correct
      (50mm max-error, above, and 50mm average-error, below) and
    visualize the median error using the colored overlays from
    Fig.~\ref{fig:thresholds}. We account for sensor specific noise artifacts using established
      techniques~\cite{Camplani2012}. Please refer
    to the text for more details. } 
\label{table:generalization}
\end{table}

\insertICLPerfPlots

We now report our experimental results, comparing datasets and
methods.
We first address the ``state of the problem'': what
aspects of the problem have been solved, and what remain open research
questions? We conclude by discussing the specific lessons we
learned and suggesting directions for future systems.

\paragraph{Mostly-solved (distinct poses):}
Fig.~\ref{fig:ICL-quant} shows that hand pose estimation is mostly
solved on datasets of uncluttered scenes where hands face
 the camera (\ie \ ICL). Deep models, decision forests, and NN all
perform quite well, both in terms of articulated pose estimation (85\%
of frames are within 50mm max-error) and hand
detection (100\% are within 100mm max-error). Surprisingly, NN
outperforms decision forests by a bit. However, when NN is trained on
other datasets with larger pose variation, performance is considerably
worse.
This suggests that the test poses remarkably resemble the training
poses. 
But, this may be reasonable
for applications targeting sufficiently distinct poses from a finite vocabulary (\eg, a gaming
interface). These results suggest that
{\em the state-of-the-art accurately predicts distinct poses}
(\ie 50 mm apart)
{\em in uncluttered scenes.}

\paragraph{Major progress (unconstrained poses): } The NYU testset still considers isolated hands, but includes a wider range of poses, viewpoints, and subjects compared to ICL (see Fig.~\ref{fig:Dataset-MDS}). 
Fig.~\ref{fig:NYU-quant} reveals that deep models perform the best for both articulated pose estimation (96\% accuracy) and hand detection (100\% accuracy). While
decision forests struggle with the added variation in pose
and viewpoint, NN still does quite well. In fact, when
measured with average (rather than max) error, NN nearly
matches the performance of~\cite{tompson14tog}.
This suggests that exemplars get most, but not
all fingers, correct
\ifIsJrn
\begin{figure}
  \begin{centering}
        \includegraphics[width=.45\columnwidth]{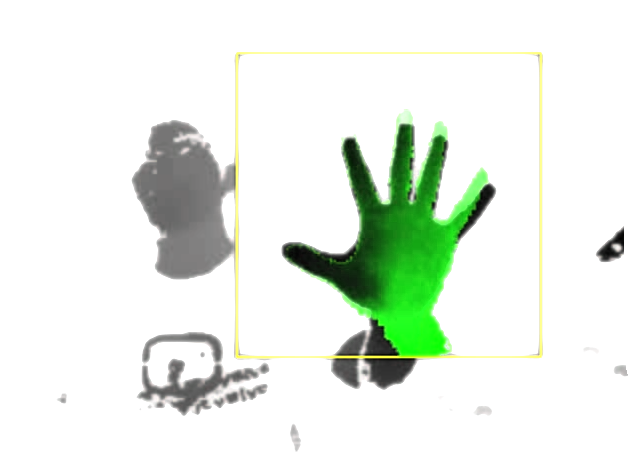} 
    \includegraphics[width=.45\columnwidth]{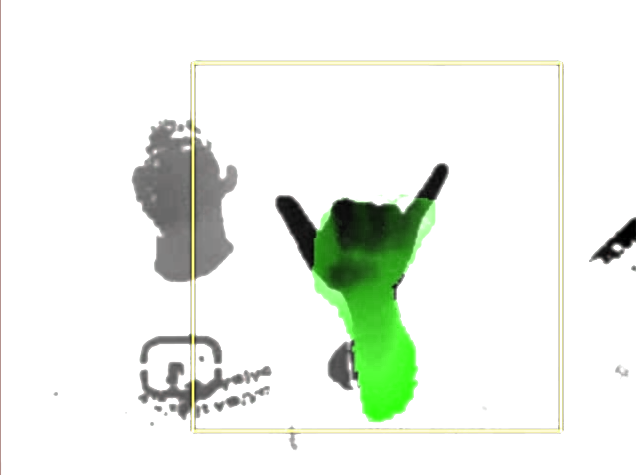}
    \label{fig:minvsmax}
    \caption{\textbf{Min vs max error:} Compared to state-of-the-art, our 1-NN baseline
      often does relatively better under the average-error criterion
      than under the max-error criterion. 
      When it can find (nearly) an exact match between training and
      test data (left) it obtains very low error. However, it does not generalize well to unseen
      poses (right). When presented with a new pose it will often
      place some fingers perfectly but others totally wrong. The
      result is a reasonable mean error but a high max error. }
  \end{centering} 
\end{figure}
(see Fig.~\ref{fig:minvsmax}).
\fi
\ifIsJrn
Overall, {\em we see noticeable progress on
unconstrained pose estimation since 2007}~\cite{Erol2007}.
\fi

\ifIsJrn
\begin{figure}
  \begin{centering}
        \includegraphics[width=.45\columnwidth]{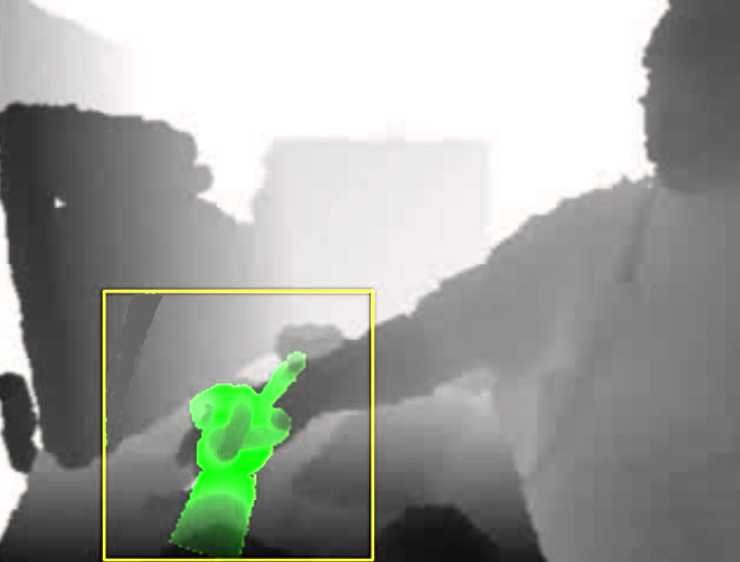} 
    \includegraphics[width=.45\columnwidth]{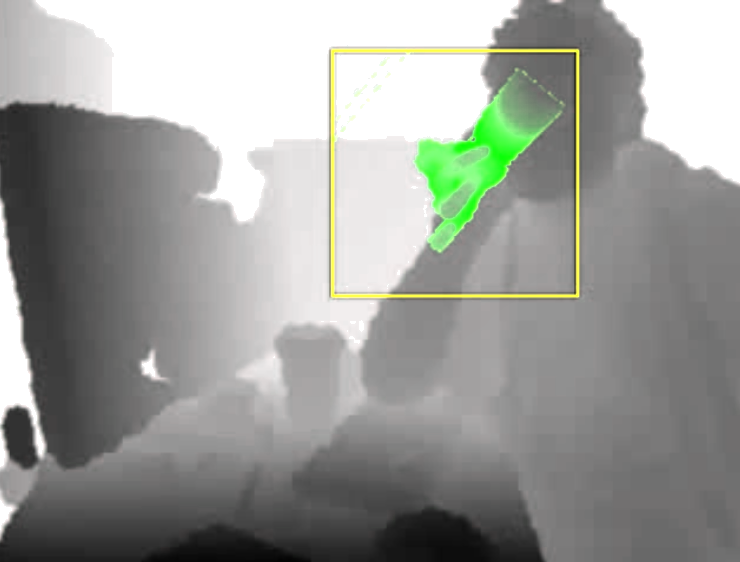}
    \label{fig:malseg}
    \caption{\textbf{Complex backgrounds: } Most existing systems, including our own 1-NN baseline,
      fail when challenged with complex backgrounds which cannot be
      trivially segmented. These backgrounds significantly alter the
      features extracted and processed  and thus prevent even
      the best models from producing sensible output.  }
  \end{centering} 
\end{figure}
\fi

\inesrtNYUPerfPlots

\ifIsJrn
\ifIsJrn
\begin{figure}
  \else
  \begin{figure}
    \fi
  
\centering
\begin{tabular}{cc}
  (a) Latent Hough Detection & (c) per-pixel classification \\
  \includegraphics[width=.4\columnwidth]{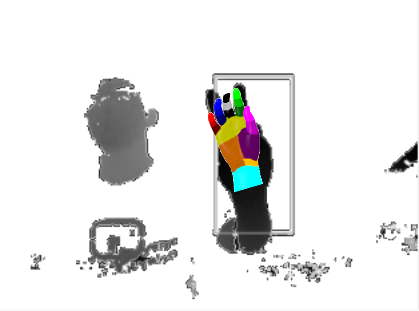} &
  \includegraphics[width=.25\columnwidth]{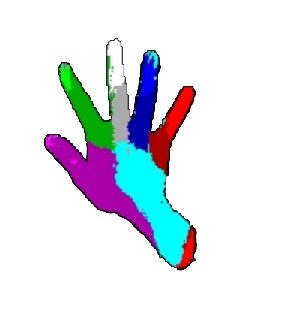} \\
  (b) Hough orientation failure & (d) hard segmentation \\
  \includegraphics[width=.4\columnwidth]{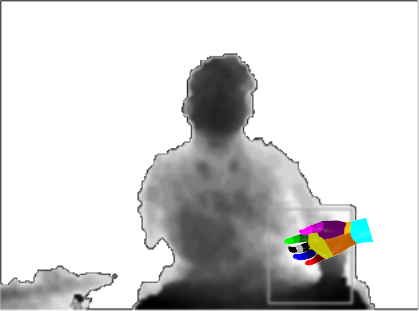} &
  \includegraphics[width=.4\columnwidth]{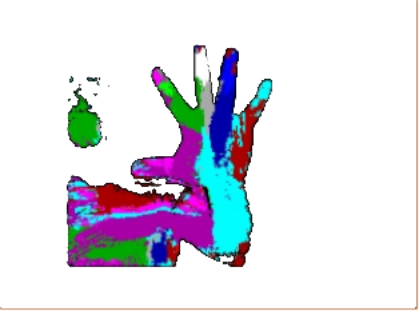} 
\end{tabular}
  \caption{
    Many approach the problem of hand pose estimation in three phases:
    (1) detect and segment (2) estimate pose (3) validate or
    refine~\cite{Xu2013,keskin2012hand,tompson14tog,tanglatent,Intel:PXC}.  
    However, when an earlier stage fails, the later stages are often
    unable to recover.
    When detection and segmentation are non-trivial, this
   becomes to root cause of many failures. 
   For example, \textbf{Hough forests}~\cite{Xu2013} (a) first 
   estimate the hand's location and
   orientation. They then convert to a cardinal translation and
   rotation before estimating joint locations. (b) When this first stage
   fails, the second stage cannot recover. 
   (c) Other methods assume that segmentation is solved 
   ~\cite{keskin2012hand,Couprie2013}, (d) when background clutter is inadvertently
   included by the hand segmenter, the finger pose estimator is prone
   to spurious outputs. 
 }
  \label{fig:NYU-hough}

\end{figure}

 \fi

\paragraph{Unsolved (low-res, objects, occlusions, clutter):}
\ifIsJrn
When considering datasets (Fig.~\ref{fig:our-quant} and~\ref{fig:ego-quant}) with distant (low-res) hands and background clutter
due to objects or interacting surfaces (Fig.~\ref{fig:malseg}), results are
significantly worse.
\fi
Note that many applications~\cite{shotton2013real} often demand hands to lie at distances greater than 750mm. For such scenes, hand detection is still a
challenge. Scanning window approaches (such as our
NN baseline) tend to outperform multistage pipelines~\cite{keskin2012hand,Couprie2013}, which may make an unrecoverable
error in the first (detection and segmentation) stage.
\ifIsJrn
We show some illustrative examples in Fig.~\ref{fig:NYU-hough}.
\fi
However,
overall performance is still lacking, particularly when compared to
human performance.
Though interestingly, human (annotator) accuracy also degrades for
low-resolution hands far away from the camera (Fig.~\ref{fig:our-quant}).
Our results suggest that {\em scenes of in-the-wild hand activity are still
beyond the reach of the state-of-the-art}.

\insertOurPerfPlots
\insertEgoPerfPlots

\ifIsJrn
\begin{figure}
\centering
    \includegraphics[width=.9\columnwidth]{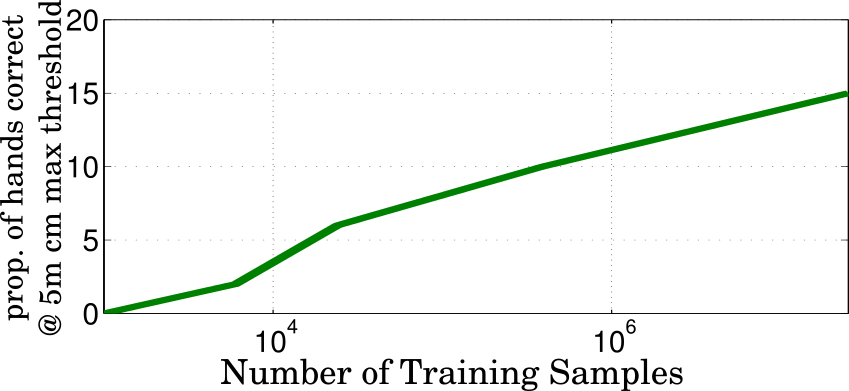}
  \caption{\textbf{Synthetic data vs. accuracy}: Synthetic
  training set  size impacts               
  performance on our test testset. Performance grows logarithmically with the
  dataset size. Synthesis is theoretically unlimited, but practically
  becomes unattractively slow. 
\label{fig:dataVsAcc}}
\end{figure}
\fi

\paragraph{Training data:} We use our NN-baseline to analyze the effect of training data in Table \ref{table:generalization}.
Our NN model performed better using the NYU training set~\cite{tompson14tog} (consisting of real data automatically labeled with a geometrically-fit 3D CAD
model) than with the libhand training set.
While performance increases by enlarging the synthetic
training set (Fig.~\ref{fig:dataVsAcc}), this quickly becomes
intractable. 
This reflects the difficulty in using synthetic data: one must carefully
model priors~\cite{Oberweger}, sensor noise,
~\cite{gupta2014learning} and hand shape variations between
users~\cite{taylor2014user}. 
Moreover, in some cases, the variation in the performance of NN (dependent on the particular training set) exceeded the
variation between model architectures (decision forests versus deep models) -
Fig.~\ref{fig:ICL-quant}. Our results suggest the diversity and
realism of the \emph{training set is as important than the model form
learned from it}. 

\paragraph{NN vs Deep models:} Overall, our 1-NN baseline proved
to be suprisingly strong, outperforming or matching the performance of
most prior systems. This holds true even for moderately-sized training
sets with tens of thousands of examples, suggesting that much prior work essentially
memorizes training examples. One contribution of our analysis
is the notion that {\em NN-exemplars provides a vital baseline for
understanding the behavior of a proposed system in relation to its
training set.}  In fact, DeepJoint~\cite{tompson14tog} and
DeepPrior~\cite{Oberweger} were the sole approaches to significantly
outperform 1-NN (Figs.~\ref{fig:ICL-quant} and \ref{fig:NYU-quant}). This
indicates that deep architectures generalize well to novel test
poses. This may contrast with existing folk wisdom about deep models:
that the need for large training sets suggests that these models
essentially memorize. Our results indicate otherwise.

\paragraph{Conclusion:}
The past several years have shown tremendous progress regarding hand pose: training sets, testing sets, and
models. Some applications, such as gaming interfaces and sign-language recognition, appear to be well-within reach for current systems. Less than a decade ago, this was not true 
\ifIsJrn~\cite{premaratne2010human,Erol2007,Cooper2012}. \fi
 Thus, we have
made progress! But, challenges remain nonetheless. Specifically, when segmentation is hard due to
active hands or clutter, many existing methods fail. To illustrate these
realistic challenges we introduce a novel testset. We demonstrate that
realism and diversity in training sets is crucial, and can be as
important as the choice of model architecture. In terms of model
architecture, we perform a broad benchmark evaluation and find that
deep models appear particularly well-suited for pose estimation. 
Finally, we demonstrate that NN using volumetric exemplars provides a
startlingly potent baseline, providing an additional tool for analyzing both methods and datasets.

{\small
\bibliographystyle{ieee}
\insertBibliography
}

\end{document}